\documentclass[pdflatex,sn-mathphys]{sn-jnl}

\jyear{2021}%

\theoremstyle{thmstyleone}%
\theoremstyle{thmstyletwo}%

\theoremstyle{thmstylethree}%

\begin{document}

\title[Logistic-ELM]{Logistic-ELM: A Novel Fault Diagnosis Method for Rolling Bearings}








\author*{\fnm{Zhenhua} \sur{Tan}}\email{tanzh@mail.neu.edu.cn}

\author{\fnm{Jingyu} \sur{Ning}}\email{ningjy@mail.neu.edu.cn}

\author{\fnm{Kai} \sur{Peng}}\email{pengkai@stumail.neu.edu.cn}

\author{\fnm{Zhenche} \sur{Xia}}\email{xiazc@stumail.neu.edu.cn}

\author{\fnm{Danke} \sur{Wu}}\email{wudk@stumail.neu.edu.cn}

\affil*{\orgdiv{Software College}, \orgname{Northeastern University}, \orgaddress{\street{Heping}, \city{Shenyang}, \postcode{110819}, \state{Liaoning}, \country{China}}}


\abstract{The fault diagnosis of rolling bearings is a critical technique to realize predictive maintenance for mechanical condition monitoring. In real industrial systems, the main challenges for the fault diagnosis of rolling bearings pertain to the accuracy and real-time requirements. Most existing methods focus on ensuring the accuracy, and the real-time requirement is often neglected. In this paper, considering both requirements, we propose a novel fast fault diagnosis method for rolling bearings, based on extreme learning machine (ELM) and logistic mapping, named logistic-ELM. First, we identify 14 kinds of time-domain features from the original vibration signals according to mechanical vibration principles and adopt the sequential forward selection (SFS) strategy to select optimal features from them to ensure the basic predictive accuracy and efficiency. Next, we propose the logistic-ELM for fast fault classification, where the biases in ELM are omitted and the random input weights are replaced by the chaotic logistic mapping sequence which involves a higher uncorrelation to obtain more accurate results with fewer hidden neurons. We conduct extensive experiments on the rolling bearing vibration signal dataset of the Case Western Reserve University (CWRU) Bearing Data Centre. The experimental results show that the proposed approach outperforms existing state-of-the-art comparison methods in terms of the predictive accuracy, and the highest accuracy is 100\% in seven separate sub data environments. Moreover, in terms of the runtime cost, the experimental results indicate that the proposed logistic-ELM can predict the fault in 40 ms with a high accuracy, up to 21–1858 times more rapidly than existing methods based on SVM, CNN and multi-scale entropy. Other experiments of fault diagnosis of the rolling bearings under four different loads, also indicate that the logistic-ELM can adapt to different operation conditions with high efficiency. The relevant code is publicly available at https://github.com/TAN-OpenLab/logistic-ELM.}

\keywords{Fault diagnosis, Rolling bearing, Extreme learning machine, Logistic mapping, Machine learning}



\maketitle

\section{Introduction}
Rolling bearings are common components of rotating machinery, which is widely used in the electric power, metallurgical, petrochemical, machinery manufacturing, aerospace domain and other fields. Due to the long-term operation of rolling bearings in harsh operating environments, the probability of fault occurrence is extremely high. For example, in a wind turbine gearbox, bearing failure accounts for 76\% of all failures \cite{1}. Rolling bearing faults often trigger a series of chain reactions, ranging from the interruption of the production process, which leads to economic loss, to catastrophic accidents, which may lead to casualties. Therefore, the fault diagnosis of rolling bearings is of significance to both the industry and academia. .

\subsection{Existing Situation}
From 1960s, researchers dedicated to develop fault diagnosis of rolling bearings through different mechanisms such as vibration, static electricity, temperature and ferro graphs. Initially, people diagnosed rolling bearing faults by developing simple instruments, such as \cite{2,3,4,5}, which were usually based on some specific physical principles but with lower accuracy. With the rapid development of the signal analysis theory, especially after the emergence of the fast Fourier transform \cite{6}, people adopted a variety of spectrum analysers to diagnose bearing faults. Schemes based on feature extraction from time-domain or frequency-domain improved the performance of fault diagnosis. Time-domain feature analysis methods usually determine the possible running states of rolling bearings by calculating and analyzing parameters and indexes of the vibration signals with various time-domain feature parameters, such as \cite{7,8,9,10,11,12,13}, while frequency-domain feature analysis methods focus on separating or strengthening the frequency components of the fault signals, such as \cite{14,15,16,17,18,19,20}, generally with higher accuracies than those of time-domain based methods. The time-frequency analysis methods combined the two to form a joint function, could describe the non-linear and non-stationary dynamic signals of complex mechanical equipment, such as \cite{21,22}. Recently, with the development of machine learning and deep learning, the accuracy of fault diagnosis of rolling bearings obtained great improvement, such as support vector machine based methods \cite{23,24}, Bayesian classifier \cite{25}, and neural network based algorithms \cite{27,28,29,30,37,38}. More details are described in Section \ref{section2}.

\subsection{Motivation}
In real industrial systems, the main challenges for the fault diagnosis of rolling bearings pertain to the accuracy and real-time requirements. However, most existing methods focus on ensuring the accuracy, and the real-time requirement is often neglected. Traditional research based on time- or frequency- domain usually focused on examining the properties of a certain indicator and returned an approximate diagnosis but not real-time classification of diagnosis. Machine learning or deep learning based methods usually directly input the vibration signals into a classifier but didn’t consider relevance of and redundancy in the feature indicators, owing to which, the classification may need more time cost before the diagnosis output. Furthermore, in practical applications, fault diagnosis is accompanied through mechanical work where the response speed must be sufficiently high, so we need a more accurate and faster classification for fault diagnosis of rolling bearings.
\subsection{Solution and Contribution}
To improve the above-mentioned, this paper proposes a novel fault diagnosis method for rolling bearings, known as the logistic-ELM. We select the extreme learning machine (ELM) as the baseline classifier in our method. To obtain the information contained in the vibration signals as fully as possible and ensure the accuracy, we first model the original vibration signals and extract a group of key features in the field of mechanical vibration based on the principle of mechanical signals, then utilize sequential forward selection (SFS) to reduce the information redundancy may exist among the features and select the most valuable features for the fault diagnosis. To obtain a more stable predictive accuracy, we use pseudo-random sequence generated by a chaotic Logistic Mapping \cite{36} to replace the random input weight matrix in ELM. Using the combination of the logistic mapping and the ELM algorithm, a pseudo-random sequence is generated through logistic mapping, and the values in the sequence are arranged into a matrix as the input weight matrix of the ELM. And the bias matrix of the ELM is set as zero. Under the same diagnosis environment, the input weight matrix needs to be generated only once, and the input weights can be reused in the subsequent diagnosis. The main contributions of this paper are as follows:
\begin{enumerate}
    \item We consider 14 kinds of physical mechanical principles to extract time-domain features from vibration signals, which can be calculated fast during fault diagnosis. Multiple signal features are jointly considered in the processes of fault diagnosis, and the advantages of each feature are cross fused to ensure the predictive accuracy and efficiency. 
    \item The proposed logistic-ELM fully exploits the internal un-correlation of the chaotic logistic mapping, and enhance the difference between the hidden neurons by generating input weight matrix through logistic mapping, further ensuring the runtime efficiency and predictive accuracy in fault diagnosis of rolling bearings.
    \item Experimental results show that the proposed logistic-ELM outperforms existing state-of-the-art comparison methods in terms of the predictive accuracy, and can realize the fault diagnosis with a reasonable predictive accuracy in 40 ms, up to 21–1858 times more rapidly than existing comparisons.
\end{enumerate}

The remaining article is organized as follows. The related work and basic principle of the ELM is briefly described in Section \ref{section2}. The design of the proposed method is described in Section \ref{section3}. We use the rolling bearing vibration signal data set of the Case Western Reserve University to perform a simulation to verify and analyse the proposed method, as described in Section \ref{section4}. The concluding remarks are presented in Section \ref{section5}.

\section{Related Work and Preliminary}
\label{section2}
\subsection{Related Work}
Research on the fault diagnosis of rolling bearings began in the 1960s \cite{2,3}. Initially, people diagnosed bearing faults by using simple instruments such as listening sticks. In 1971, a Swedish instrument company first developed the shock pulse measurement (SPM) technique to detect bearing faults \cite{4}. The SPM is based on the principle that the bearing failure impact can induce the resonance of the system, and bearing faults are diagnosed according to the magnitude of the impact pulse. However, this technique can only identify whether the bearing is faulty or not but the fault types cannot be classified, and the technique exhibits a weak anti-interference ability and low stability. In general, the identification of a fault occurrence in a rolling bearing and classification of the fault type can be realized considering the feature frequency of the faults in vibration signal spectra. In 1974, \cite{5} presented a patent for a resonance demodulation analysis system. This system processes the fault signal considering the resonance and by using a band-pass filter, which enhances the signal-to-noise ratio of the signal. Moreover, the accuracy of the diagnosis result is increased through envelope processing. However, a certain experience is necessary to determine the centre frequency and bandwidth of the filter, and the efficiency of diagnosis is low. In recent years, with the rapid advancement of computer technologies, various new signal processing methods have emerged. Many scholars have successively proposed computer-centric rolling bearing fault diagnosis methods, and a fault diagnosis scheme based on feature extraction and fault classification has been established.

Feature extraction approaches can be primarily divided into three categories, namely, time-domain, frequency-domain, and time-frequency domain feature extraction. The time-domain feature extraction is based on various time-domain feature parameters and performance indexes of the vibration signals. By analysing the parameters and indexes, the fault of the equipment can be effectively determined in an intuitive and easy to understand manner \cite{7}. In general, traditional time-domain features can be divided into dimensional and non-dimensional parameter indexes, which are easy to calculate and can reflect the working state of the bearings to a certain extent. The traditional research is focused on examining the properties of a single parameter, and fault diagnosis schemes combining multiple traditional time-domain features remain to be developed. In addition to traditional time-domain feature extraction methods, the information entropy theory has been applied for time-domain feature extraction \cite{8,9,10,11,12,13,41}. However, the computation time of the information entropy algorithm is usually large because of its complexity.

In contrast to the time-domain analysis, frequency-domain analysis is focused on separating or strengthening the frequency components of the fault signal. This analysis technique is widely adopted in mechanical fault diagnosis because the distribution of the frequency components of the signal in the spectrum can be obtained through the fast Fourier transform, which is more intuitive than the time-domain waveform. The Fourier transform performs statistical averaging on the signal in the time domain by integral transformation, thereby smoothening the non-stationary components in the signal. Although this approach can reflect the frequency information in the signal, it involves certain limitations as the change in the signal over time is not clarified. At present, the commonly used frequency-domain feature extraction methods include power spectrum analysis, envelope spectrum analysis \cite{14,15}, cepstrum analysis \cite{16,17}, high-order spectrum analysis \cite{18} and spectral kurtosis analysis \cite{19,20}.

In the time-frequency analysis, the time- and frequency-domains are combined to form a joint function to describe the non-linear and non-stationary dynamic signals of complex mechanical equipment. Common time-frequency analysis methods include the Wigner Ville distribution (WVD), short time Fourier transform (STFT), and wavelet transform (WT). Many scholars proposed adaptive decomposition methods by studying the laws and features of complex signals. In 1998,  \cite{21} proposed a method to analyse a non-stationary signal, named empirical mode decomposition (EMD). Moreover, the authors proposed a popular time-frequency analysis method known as the Hilbert-Huang transform, based on the Hilbert transform. In 2005, \cite{22} proposed the local mean decomposition method (LMD). The LMD can decompose a complex multi-component signal into several product functions (PF components) and a residual component. Each PF component consists of a product of an envelope signal and a pure FM signal. Compared with the EMD, the LMD method involves prompt convergence. Moreover, the latter approach can effectively reduce the errors generated in the iteration process and enhance the decomposition accuracy.

Fault classification are usually based on machine learning, and three types of algorithms are widely used. The first type pertains to the support vector machine \cite{23}. For example, \cite{24} proposed a fault diagnosis algorithm in 2011, in which, the feature vectors of the vibration signals are extracted through the wavelet transform, and fault classification is conducted using the SVM algorithm. The second type pertains to the Bayesian classifier \cite{25}, which is easy to understand and exhibits a high learning efficiency. However, the algorithm is implemented assuming the independence and normality of the independent and continuous variables, respectively, which may affect the accuracy of the algorithm to a certain extent. The third type pertains to neural-network-based algorithms, including the BP neural network \cite{26}, convolutional neural network \cite{27,40}, and extreme learning machine \cite{28}. For example, the fault diagnosis algorithm proposed by \cite{29} in 2020 involved 2000 input neurons and a hidden layer containing 10 neurons. Reference \cite{30} calculated the time-frequency features of vibration signals by realizing the continuous wavelet transform (CWT) of the complex Morlet wavelet and performing jointed time-frequency analysis (JTFA). Next, the CNN input was obtained through normalization, and the CNN was trained using a time-frequency map with labels. Finally, the trained model was used for fault diagnosis. The HGSA-ELM algorithm \cite{31} proposed by Luo M et al. in 2016 optimized the input weights and bias of the ELM through the real-valued gravitational search algorithm (RGSA), and the binary-valued of the GSA (BGSA) was used to select the valuable features from a compound feature set. Three types of fault features, specifically, time and frequency features, energy features and singular value features, were extracted to compose the compound feature set by applying ensemble empirical mode decomposition (EEMD).

\subsection{Preliminary: ELM}
\begin{figure}[!t]
\centering
\includegraphics[width=3in]{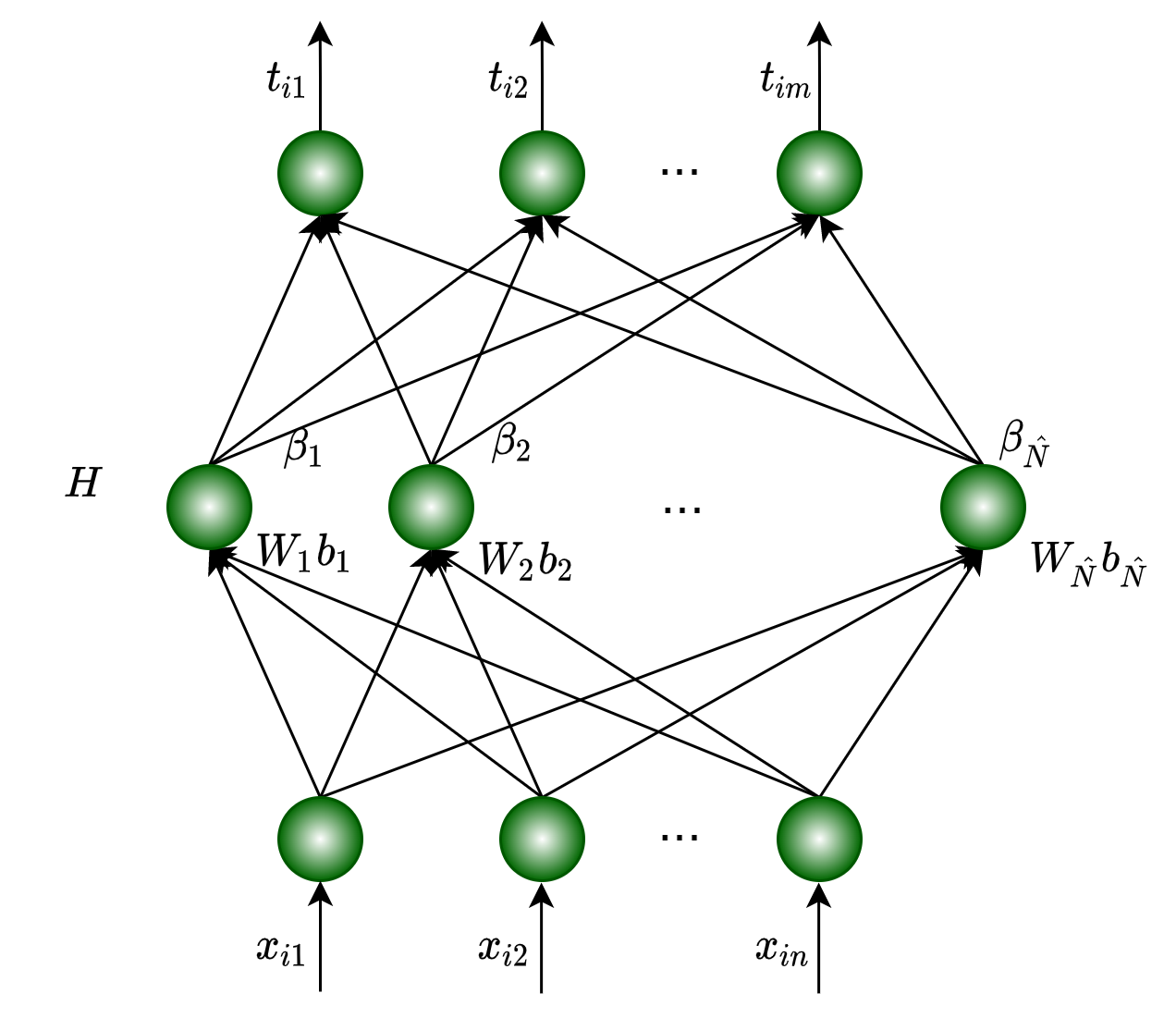}
\caption{Framework of the ELM.}
\label{fig1}
\end{figure}

The extreme learning machine (ELM) \cite{28} algorithm can be used for a single hidden layer feedforward neural network (SLFN), as shown in Fig.~\ref{fig1}. 

Traditional feedforward neural networks involve notable limitations, such as a low speed, presence of local minimum, improper learning rate and overfitting. The ELM randomly generates the connection weight between the input and hidden layers, and this weight does not need to be adjusted during the training process. Only the number of hidden layer neurons must be set to obtain the unique optimal solution. Compared with the previous traditional training methods, the ELM exhibits a higher learning speed and generalization performance. 

For $N$ arbitrary distinct samples $(x_i,t_i)$, where $x_{i}=\left[x_{i 1}, x_{i 2}, \cdots , x_{i n}\right]^{T} \in R^{n}$ and $t_{i}=\left[t_{i 1}, t_{i 2}, \cdots , t_{i m}\right]^{T} \in R^{m}$, standard SLFNs with $\tilde{N}$ hidden neurons and activation function $g(x)$ can be mathematically modelled as
\begin{equation}
    \sum_{i=1}^{N} \beta_{i} g\left(w_{i} \cdot x_{j}+b_{i}\right)=o_{j}, j=1, \cdots, N
\end{equation}
where $w_{i}=\left[w_{i, 1}, w_{i, 2}, \cdots w_{i, n}\right]^{T}$, is the weight vector connecting the $i$th hidden neuron and input neurons, and $b_i$ is the threshold of the $i$th hidden neuron. $w_i\cdot x_j$ denotes the inner product of $w_i$ and $x_j$.

The standard SLFNs with $\tilde{N}$ hidden neurons and activation function $g(x)$ can approximate the $N$ samples with zero error, specifically, $\sum_{j=1}^{N}\left\|o_{j}-t_{j}\right\|=0$. Consequently, there exist $\beta_i$, $w_i$ and $b_i$ such that
\begin{equation}
    \sum_{i=1}^{\widetilde{N}} \beta_{i} g\left(w_{i} \cdot x_{j}+b_{i}\right)=t_{j}, j=1, \cdots, N
\end{equation}
$N$ can be compactly expressed as
\begin{equation}
    H\beta=T
\end{equation}
where
\begin{equation}
    H=\left[\begin{array}{ccc}
g\left(w_{1} \cdot x_{1}+b_{1}\right) & \cdots & g\left(w_{\widetilde{N}} \cdot x_{1}+b_{\widetilde{N}}\right) \\
\vdots & \ddots & \vdots \\
g\left(w_{1} \cdot x_{N}+b_{1}\right) & \cdots & g\left(w_{\widetilde{N}} \cdot x_{N}+b_{\widetilde{N}}\right)
\end{array}\right]_{N \times \widetilde{N}}
\end{equation}

\begin{equation}
    \beta=\left[\begin{array}{c}
\beta_{1}^{T} \\
\vdots \\
\beta_{L}^{T}
\end{array}\right]_{\widetilde{N} \times m} \quad T=\left[\begin{array}{c}
t_{1}^{T} \\
\vdots \\
t_{N}^{T}
\end{array}\right]_{N \times m}
\end{equation}

Identifying the specific $\hat{w_i}$, $\hat{b_i}$ and $\hat{\beta_i}(i=1,...,\tilde{N})$ is equivalent to minimizing the cost function
\begin{equation}
    E=\sum_{j=1}^{N}\left(\sum_{i=1}^{\widetilde{N}} \beta_{i} g\left(w_{i} \cdot x_{j}+b_{i}\right)-t_{j}\right)^{2}
\end{equation}

The input weights and hidden layer biases of the SLFNs do not need be adjusted and can be arbitrarily assigned. For fixed input weights $w_i$ and hidden layer biases $b_i$, training an SLFN is equivalent to finding a least-squares solution $\hat{\beta_i}$ of the linear system $H\beta=T$. The smallest norm least-squares solution of the above linear system is

\begin{equation}
    \hat{\beta}=H^{\dagger} T
\end{equation}
where $H^{\dagger}$ is the Moore–Penrose generalized inverse of matrix $H$.
\section{Proposed method}
\label{section3}
\begin{figure}[!t]
\centering
\includegraphics[width=4.5in]{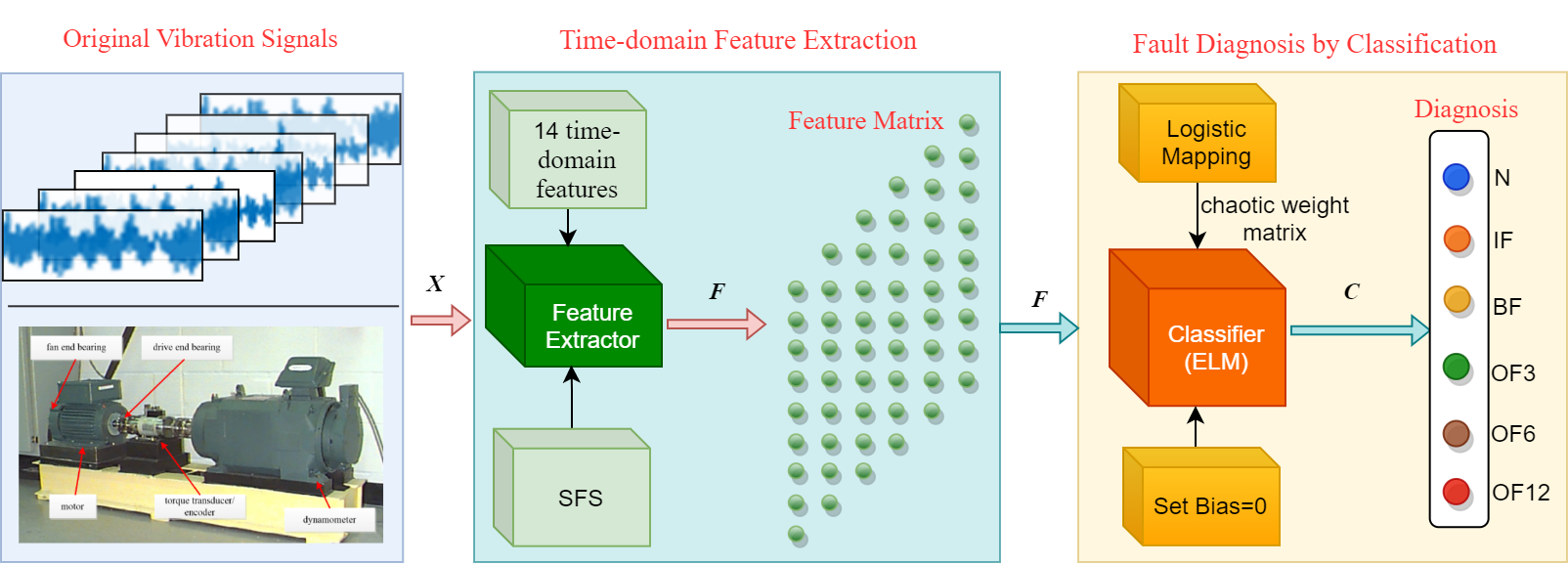}
\caption{Framework of logistic-ELM.}
\label{fig2}
\end{figure}

In this section, we describe the design of proposed logistic-ELM. Let $D:\\ \left\{x_{i}, y_{i}\right\}_{i=1}^{N}$ indicate distinct samples, where $x_{i}=\left[x_{i 1}, x_{i 2}, \cdots , x_{i n}\right]^{T} \in R^{n}$ is a n-dimension vibration signal, with corresponding fault label $y_i\in \{1,...,m\}$. For simplicity, let $X=\left[x_{1}, \cdots x_{N}\right]^{T} \in R^{N * n}$ represent vibration signals, and $Y=\left[y_{1}, \cdots, y_{N}\right]^{T} \in R^{N}$ denote the fault labels of $X$. Let $c_i\in {1,...,m}$ be the classification result of $x_i$ ,which is the factual output of the classifier. $C=\left[c_{1}, \cdots, c_{N}\right]^{T} \in R^{N}$ represents the classification result of $X$. Let $F=\left[f_{1}, \cdots, f_{N}\right]^{T} \in R^{N * K}$ be the feature matrix of $X$, where $f_i$ is the feature vector of $x_i$. The framework of the proposed method is shown in Fig.~\ref{fig2}. Based on the original vibration signals $X$, the feature extractor generates the feature matrix which is the input of the classifier ELM. 

Next, we will introduce three important components of logistic-ELM, including feature extractor, logistic-ELM classifier, and pseudo-random sequence generator by Logistic Mapping.

\subsection{Component-1: logistic-ELM Classifier}\label{sec3.1}
\begin{figure}[!t]
\centering
\includegraphics[width=4.5in]{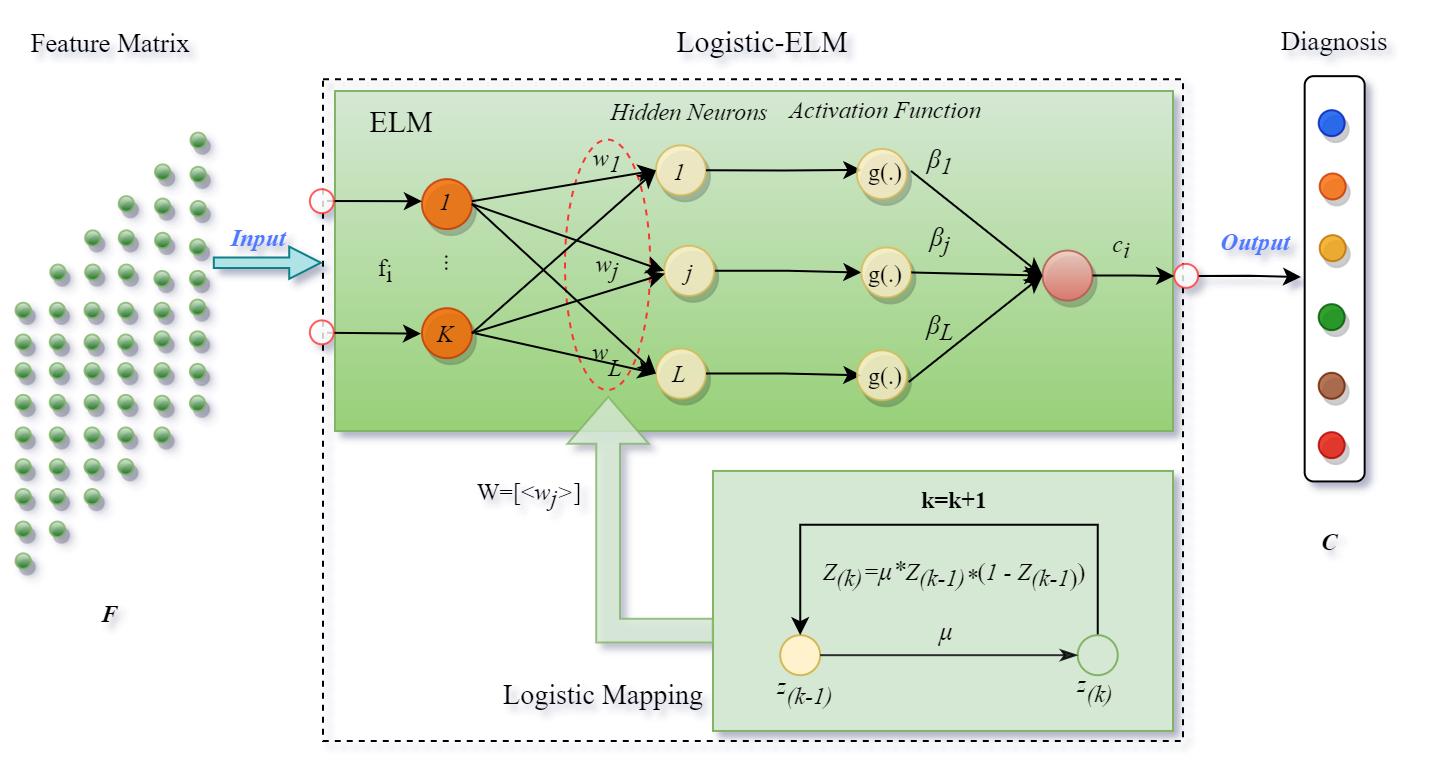}
\caption{Logistic-ELM Classifier.}
\label{fig3}
\end{figure}

The proposed logistic-ELM classifier is the core component of the proposed method, as shown in Fig.~\ref{fig3}. The purpose of this component is to classify faults based on the feature matrix $F$. Based on the original ELM, we generate the input weights through logistic mapping and omit the biases b of the ELM with $L$ hidden neurons and activation function $g(\cdot)$, the logistic-ELM can be written as
\begin{equation}
    \sum_{j=1}^{L} \beta_{j} g\left(w_{j} \cdot f_{i}\right)=c_{i}, i=1, \cdots, N
\end{equation}
where $w_{j}=\left[w_{j 1}, w_{j 2}, \cdots w_{j K}\right]^{T}$ is the weight vector connecting the input feature $f_i$ and $j$th hidden neuron, $\beta_{j}=\left[\beta_{j 1}, \beta_{j 2}, \cdots, \beta_{j m}\right]^{T}$ is the weight vector connecting the $j$th hidden neuron and output $c_i$.

Let $W=[w_1,w_2,...,w_L]$ be the input weights, and $\beta=\left[\beta_{1}, \beta_{2}, \cdots, \beta_{L}\right]^{T}$ be the output weights, such that, $H\beta=C$, where

\begin{equation}
H=\left[\begin{array}{ccc}
g\left(w_{1} \cdot f_{1}\right) & \cdots & g\left(w_{L} \cdot f_{1}\right) \\
\vdots & \ddots & \vdots \\
g\left(w_{1} \cdot f_{N}\right) & \cdots & g\left(w_{L} \cdot f_{N}\right)
\end{array}\right]_{N \times L}      
\end{equation}

\begin{figure}[!t]
\centering
\includegraphics[width=4.5in]{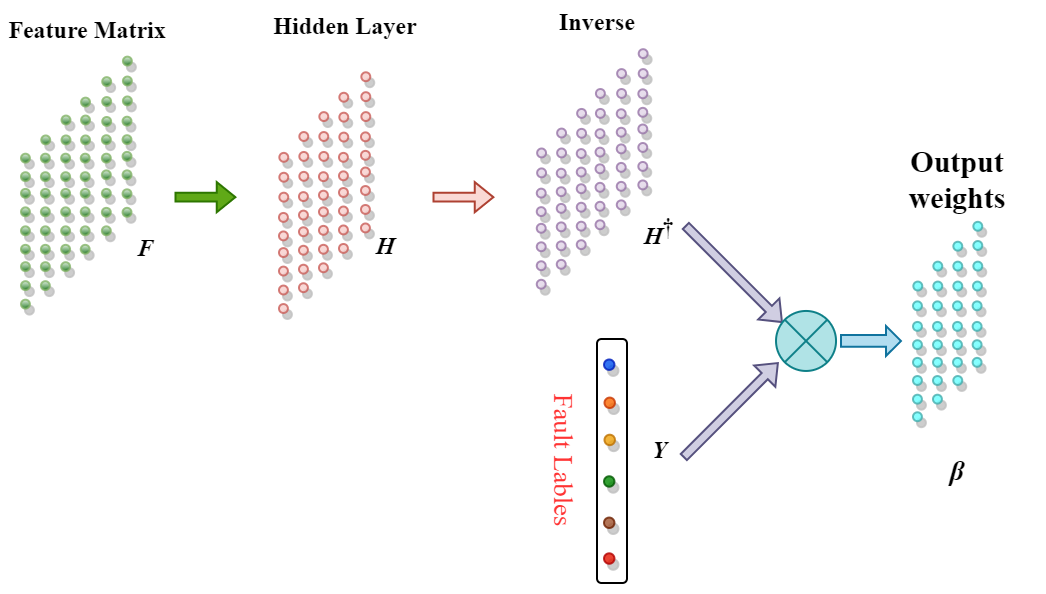}
\caption{Training process of $\beta$.}
\label{fig4}
\end{figure}

In the ideal situation, the factual output of the classifier is the same as the fault label, specifically, $C=Y$. Consequently, $H\beta=Y$, and $\hat{\beta}=H^{\dagger} Y$. The process is illustrated in Fig.~\ref{fig4}. 
\subsection{Component-2: Pseudo-random Sequence Generator for Input Weights Based on Chaotic Logistic Mapping}
As described previously, the ELM can randomly initialize the input weights and hidden layer biases to obtain the corresponding output weights, and thus, this method can be adapted to different applications. But random initialization may lead to a certain degree of uncertainty in the classification result. Therefore, we introduce the pseudo-random sequence generator of Logistic Mapping for input weights of ELM. In fact, this component is also a necessary part of logistic-ELM as shown in Fig.~\ref{fig3}. Since the function of the input weights is to map the feature matrix to the hidden neurons, the correlation between the elements of the input weights matrix should be as small as possible. To this end, we generate the input weights through logistic mapping. Logistic mapping involves the stability and instability of the complete and part of the situation, respectively, and can thus be applied to generate the initial input weights. The logistic mapping can be expressed as
\begin{equation}
    z_k=\mu z_{k-1}(1-z_{k-1})
\end{equation}
where $z_1\in (0,1)$ and $\mu \in (3.56995,4)$. Consequently, $z=(z_1,z_2,...)$, and we can obtain the initial input weights W by arranging the elements in the sequence $z$. Therefore, W can be expressed as
\begin{equation}
    W=\left[\begin{array}{ccc}
Z_{1} & \cdots & Z_{K *(L-1)+1} \\
\vdots & \ddots & \vdots \\
Z_{K} & \cdots & Z_{K * L}
\end{array}\right]
\end{equation}
In this case, a high degree of irrelevance must be ensured for $W$, and thus, the biases do not need to be set. The biases can be assigned a zero value during the classification. The input weights are generated as follows.

\begin{algorithm}[h]  
  \caption{ Weight Generator $(z_1, \mu, L, K)$.}  
  \label{alg:Framwork}  
  \begin{algorithmic}[1]  
    \Require  
      $(z_1, \mu, L, K)$
    \Ensure  
      $W$  
    \For{ $k$ 2 to $K*L$ by 1}
        \State $z_k=\mu z_(k-1) (1-z_(k-1) ) $
    \EndFor
    \For{$i$ 1 to $K$ by 1}
        \For{j 1 to L by 1}
            \State $W_{ij}=z_((i-1)*L+j)$
        \EndFor
    \EndFor\\
    \Return $W$  
  \end{algorithmic}  
\end{algorithm}  
\subsection{Component-3: Feature Extractor based on the Mechanical Vibration Principle}
\begin{figure}[!t]
\centering
\includegraphics[width=4.5in]{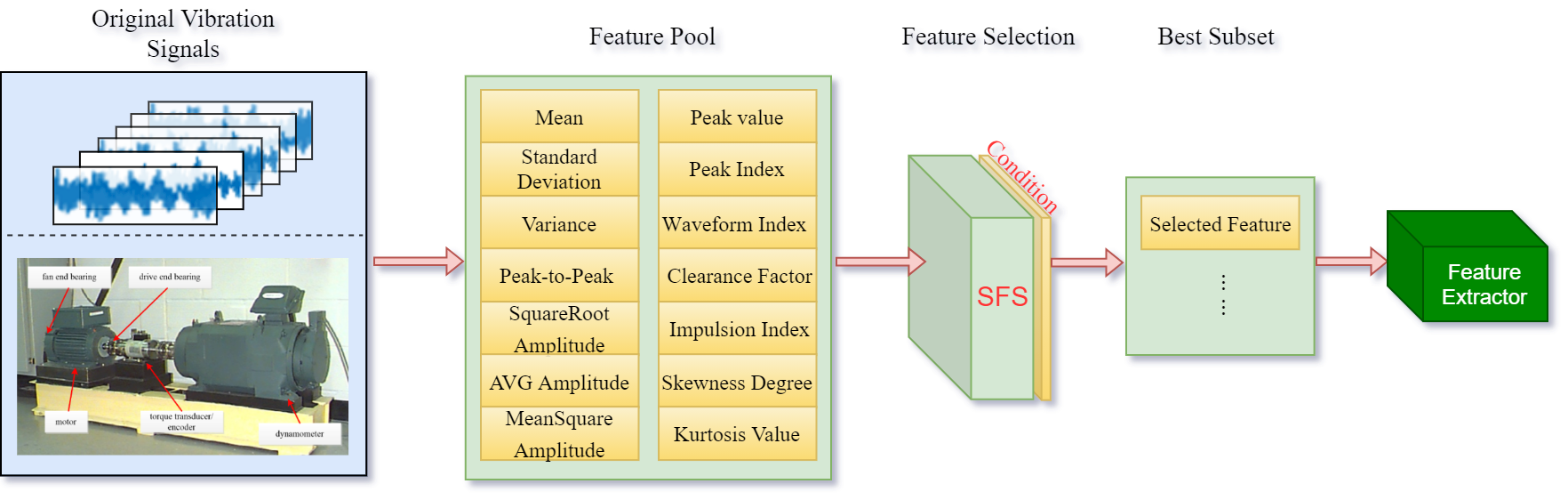}
\caption{The Process to Build the Feature Extractor.}
\label{fig5}
\end{figure}

The purpose of this component is to extract the feature matrix from vibration signals. See Fig.~\ref{fig5}.
Firstly, we extract 14 time-domain features through the mechanical principle, to express a certain situation of the vibration signals from different perspective, including Mean Value, Standard Deviation, Variance, Peak-to-Peak Value, Square Root Amplitude, Average Amplitude, Mean Square Amplitude, Peak Value, Waveform Index, Peak Index, Impulsion Index, Clearance Factor, Degree of Skewness, and Kurtosis Value. Each feature represents one characteristic of the original vibration signals, and related equations are

\begin{itemize}
    \item Mean Value: $mean\left(x_{i}\right)=\frac{1}{n} \sum_{t=1}^{n} x_{i, t}$
    \item Standard Deviation: $\sigma\left(x_{i}\right)= \sqrt{\frac{1}{n} \sum_{t=1}^{n}\left(x_{i, t}-mean\left(x_{i}\right)\right)^{2}}$
    \item Variance: $\sigma^{2}\left(x_{i}\right)=\frac{1}{n} \sum_{t=1}^{n}\left(x_{i, t}-mean\left(x_{i}\right)\right)^{2}$
    \item Peak-to-Peak Value: $Vpp\left(x_{i}\right)=\max \left(x_{i}\right)-\min \left(x_{i}\right)$
    \item Square Root Amplitude: $X r(x_{i})=(\frac{1}{n} \sum_{t=1}^{n} \sqrt{\rvert x_{i, t}\rvert})^{2}$
    \item Average Amplitude: $\rvert mean(x_{i})\rvert=\frac{1}{n} \sum_{t=1}^{n}\rvert x_{i, t}\rvert$
    \item Mean Square Amplitude: $ X r m s\left(x_{i}\right)=\sqrt{\frac{1}{n} \sum_{t=1}^{n} x_{i, t}^{2}}$
    \item Peak Value: $X p(x_{i})=\max (\rvert mean(x_{i})\rvert )$
    \item Waveform Index: $X w\left(x_{i}\right)=\frac{X r m s(x_{i})}{\mid mean(x_{i}) \mid}$
    \item Peak Index: $Ip\left(x_{i}\right)=\frac{X p\left(x_{i}\right)}{X r m s\left(x_{i}\right)}$
    \item Impulsion Index: $C f\left(x_{i}\right)=\frac{X p\left(x_{i}\right)}{mean\left(x_{i}\right)}$
    \item Clearance Factor: $Ce\left(x_{i}\right)=\frac{X r m s\left(x_{i}\right)}{mean\left(x_{i}\right)}$
    \item Degree of Skewness: $Cw(x_{i})=\frac{\frac{1}{n} \sum_{t=1}^{n}(\rvert x_{i, t}\rvert - mean \left(x_{i}\right))^{3}}{\sigma\left(x_{i}\right)^{3}}$
    \item Kurtosis Value: $Cq(x_{i})=\frac{\frac{1}{n} \sum_{t=1}^{n}(\rvert x_{i, t}\rvert-mean(x_{i}))^{4}}{\sigma(x_{i})^{4}}$
\end{itemize}
The above 14 feature models can be combined to create a feature pool $S=(s_1,...,s_{14})$ of candidate features from which the best subset $S'$ is selected by SFS procedure. Initializing the best subset as an empty set, the SFS starts the iterative processes to find appropriate features. The selection condition is of best accuracy by logistic-ELM component (described in Section \ref{sec3.1}). By comparing the classification result $C$ with the fault label $Y$, the accuracy can be calculated as
\begin{equation}
    Accuracy(C, Y)=\frac{\sum_{i=1}^{N} P\left(c_{i}=y_{i}\right)}{N} * 100 \%
\end{equation}
where $    P\left(c_{i}=y_{i}\right)=\left\{\begin{array}{ll}
1, & c_{i}=y_{i} \\
0, & c_{i} \neq y_{i}
\end{array}\right.$
Repeat the iteration until the accuracy of the best subset does not improve with the further addition of a feature. The feature extractor is built by the feature models in the best subset. Related SFS algorithm for feature selection is described as follows.

\begin{algorithm}[h]  
  \caption{SFS $(S,Y,W,\beta)$.}  
  \label{alg:SFS}  
  \begin{algorithmic}[1]  
    \Require  
      $(S,Y,W,\beta)$
    \Ensure  
      $S'$  
      
    \State	$S'=\emptyset$, $s_0=\emptyset$
    \State	$Acc=-1$, $sign = 0$, $max=0$
    \While {$Acc<max$}
    \State $Acc=max$
    \State add $s_{sign}$ to $S'$
    \State delete $s_{sign}$ from $S$
    \For{$k$ 1 to card($S$) by 1}
    \State $C=g(S'*W)*\beta$
    \If{Accuracy(C,Y) $>$ max}
    \State max= Accuracy(C,Y)
    \State sign = k
    \EndIf
    \EndFor
    \EndWhile\\
    \Return $S'$   
  \end{algorithmic}  
\end{algorithm}  

\section{Experiments and Analysis}
\label{section4}
To demonstrate the effectiveness of the logistic-ELM, we use the rolling bearing vibration signal dataset \cite{32} prepared by the Case Western Reserve University (CWRU) Bearing Data Centre to conduct experiments. This dataset is a world-recognized standard dataset for bearing fault diagnosis and contains a large number of rolling bearing vibration signals under normal and fault conditions. In this paper, we select bearing data with two kinds of fault diameter (0.007 inch and 0.021 inch) for classification. The motor load and speed are 0 hp and 1797 r/min, respectively. The sampling frequency is 12 kHz. There exist 6 fault types: normal (N), inner raceway fault (IF), ball fault (BF), and outer raceway faults located at 3 o’clock (OF3), 6 o’clock (OF6), and 12 o’clock (OF12). We select 60 original samples for each fault type, where each sample has 2048 sampling signal points. The sample description for the different fault types is presented in Table \ref{tab1}.
\begin{table}[!t]
\caption{Sample description}
\label{tab1}
\centering
\begin{tabular}{llllll}
\toprule
Fault &	Diameter&	Classification& \multicolumn{3}{l}{\#Samples}\\
\cline{4-6}
 Type&(inches) &  Label & \#Train&	\#verify	&\#Test\\
\hline
N&	0&	1	&81926	&20480&	20480\\

IF&	0.007&	2	&81926	&20480&	20480\\

IF&	0.021&	3	&81926	&20480&	20480\\

BF&	0.007&	4	&81926	&20480&	20480\\

BF&	0.021&	5	&81926	&20480&	20480\\

OF3&	0.007&	6	&81926	&20480&	20480\\

OF3	&0.021&	7	&81926	&20480&	20480\\

OF6	&0.007&	8&81926	&20480&	20480\\

OF6	&0.021&	9&81926	&20480&	20480\\

OF12&	0.007&	10&81926	&20480&	20480\\

OF12&	0.021&	11&81926	&20480&	20480\\
\hline
\end{tabular}
\footnotetext{Note: frequency=12 kHz; motor load=0 hp; speed=1797 r/min.}
\end{table}

\begin{figure}[!t]
\centering
\includegraphics[width=4in]{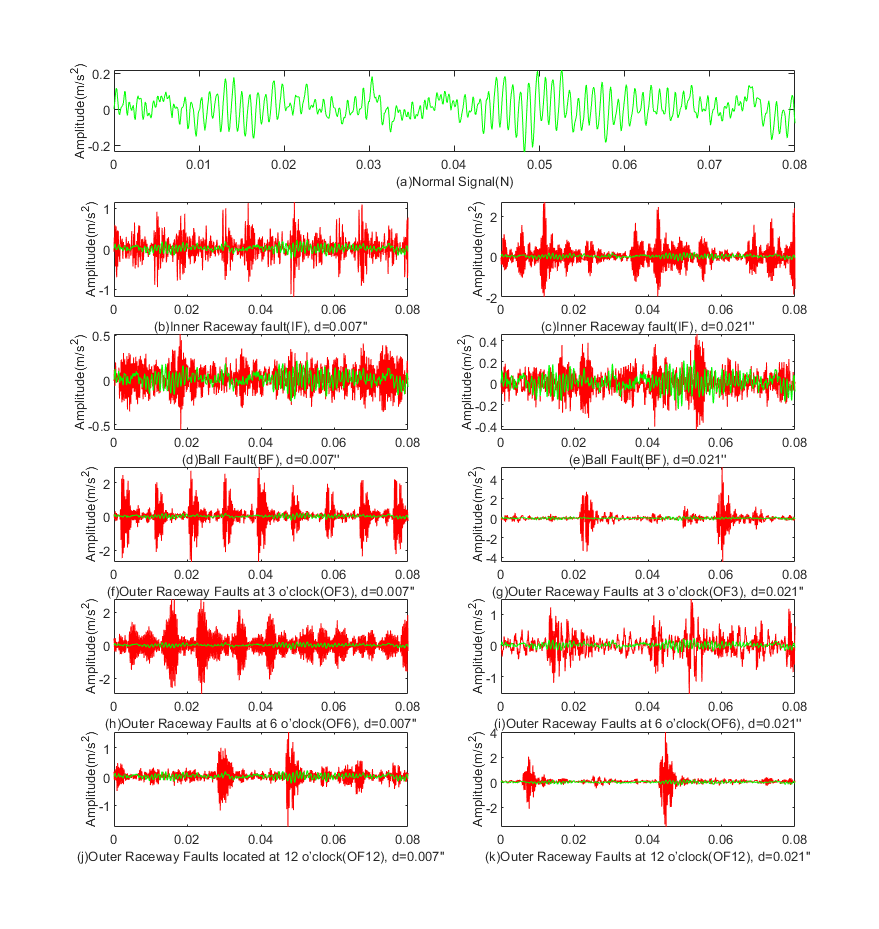}
\caption{The Process to Build the Feature Extractor.}
\label{fig6}
\end{figure}

Fig.~\ref{fig6} shows the original vibration signals of the 11 fault types, where the green and red lines indicate the normal signals and fault signals, respectively. The amplitude of the normal signal is small, as shown in Fig.~\ref{fig6}(a), and the periodicity is not notable. We observe that there are great amplitude deviations between fault signals and normal signals, as shown in Fig.~\ref{fig6}(b) to (k), and some of the fault signals generally demonstrates an obvious periodicity. As shown in Fig.~\ref{fig6}(b) and (c), we can observe periodic in inner race fault with different size. The 0.007-inch inner race fault causes the amplitude to increase by approximately 4 times, while the 0.021-inch inner race fault causes the amplitude to increase by approximately 10 times. Therefore, for inner race fault, larger diameter leads to larger amplitude. The ball fault causes the amplitude to increase by approximately 2 times, and the signal does not exhibit any notable periodicity, as shown in Fig.~\ref{fig6}(d) and (e), likely because the balls roll constantly, and the contact points with the raceway change randomly. Coupled with the effect of the lubricating oil, the fault performance is not sufficiently clear. The outer raceway faults are periodic with a large amplitude as shown in Fig.~\ref{fig6}(f) to (k). For the outer raceway faults located at 3 o’clock and 12 o’clock, larger diameter leads to larger amplitude. On the contrary, for the outer raceway faults located at 6 o’clock, larger diameter leads to smaller amplitude. Therefore, the linear correlation between fault diameter and amplitude cannot be determined.

The vibration signals of different bearing faults have different features and exhibit a certain regularity, which is the theoretical basis for fault diagnoses. However, the features of the signals cannot be obtained through only manual observation, and thus, a scientific algorithm design must be implemented.

We deploy six experiments to verify the accuracy and efficiency of proposed logistic-ELM with comparisons from multiple perspectives. Experiment 1-5 are the analysis of logistic-ELM itself, and Experiment 6 and 7 are the comparation with existing state-of-the-art methods. Experiment 1, 2, 3, 4 and 7 are performed on CWRU dataset with fault diameter d = 0.007’’ \& 0.021’’ and load = 0hp as described above, while experiment 5 is performed on CWRU dataset with fault diameter d=0.007’’ \& 0.021’’ and load = 0hp \& 1hp \& 2hp \& 3hp. Experiment 6 is performed on CWRU database with diameters identical to the comparison methods.

The purpose of each experiment is as follows. {\bfseries Experiment 1} is to select appropriate number of hidden neurons and the activation function for logistic-ELM. {\bfseries Experiment 2} is to select initial chaotic values $z_1$ and $\mu$ for Logistic Mapping of logistic-ELM.
{\bfseries Experiment 3} shows the process of selecting the features for logistic-ELM by SFS. {\bfseries Experiment 4} is about predictive accuracy comparison with original ELM. Parameters of Experiment 1-4 are set as Table \ref{tab2}. {\bfseries Experiment 5} is to verify the performance of logistic-ELM under different operating conditions. {\bfseries Experiment 6} is about predictive accuracy comparisons with existing fault diagnosis methods. {\bfseries Experiment 7} is about runtime cost comparisons with existing fault diagnosis methods.

\begin{table}[!t]
\caption{Parameters of Experiments}
\label{tab2}
\centering

\resizebox{4.5in}{!}{
\begin{tabular}{cccccc}
\toprule
Value&	Exp1&	Exp2& Exp3&\multicolumn{2}{c}{Exp4}\\
\cline{5-6}
&&&& ELM	&Logistic-ELM\\
\hline
Activation Function&	Input&	Sigmoid&	Sigmoid& Sigmoid	&Sigmoid\\

Hidden Neurons&	Input&	20&	20&	20&20\\

$\mu$&	3.9	&Input&	3.9	&- &  Selected in Exp2\\

$z_1$&	0.6&Input	&0.6	&- & Selected in Exp2\\

Data Samples&	Train, Verify&	Train, Verify&Train, Verify&	Train, Verify, Test&	Train, Verify, Test\\
\hline
\end{tabular}}
\end{table}

\subsection{Experiment 1. Select Appropriate Activation Function and the Number of Hidden Neurons for logistic-ELM}
\label{exp1}

\begin{figure}[!t]
\centering
\includegraphics[width=3.5in]{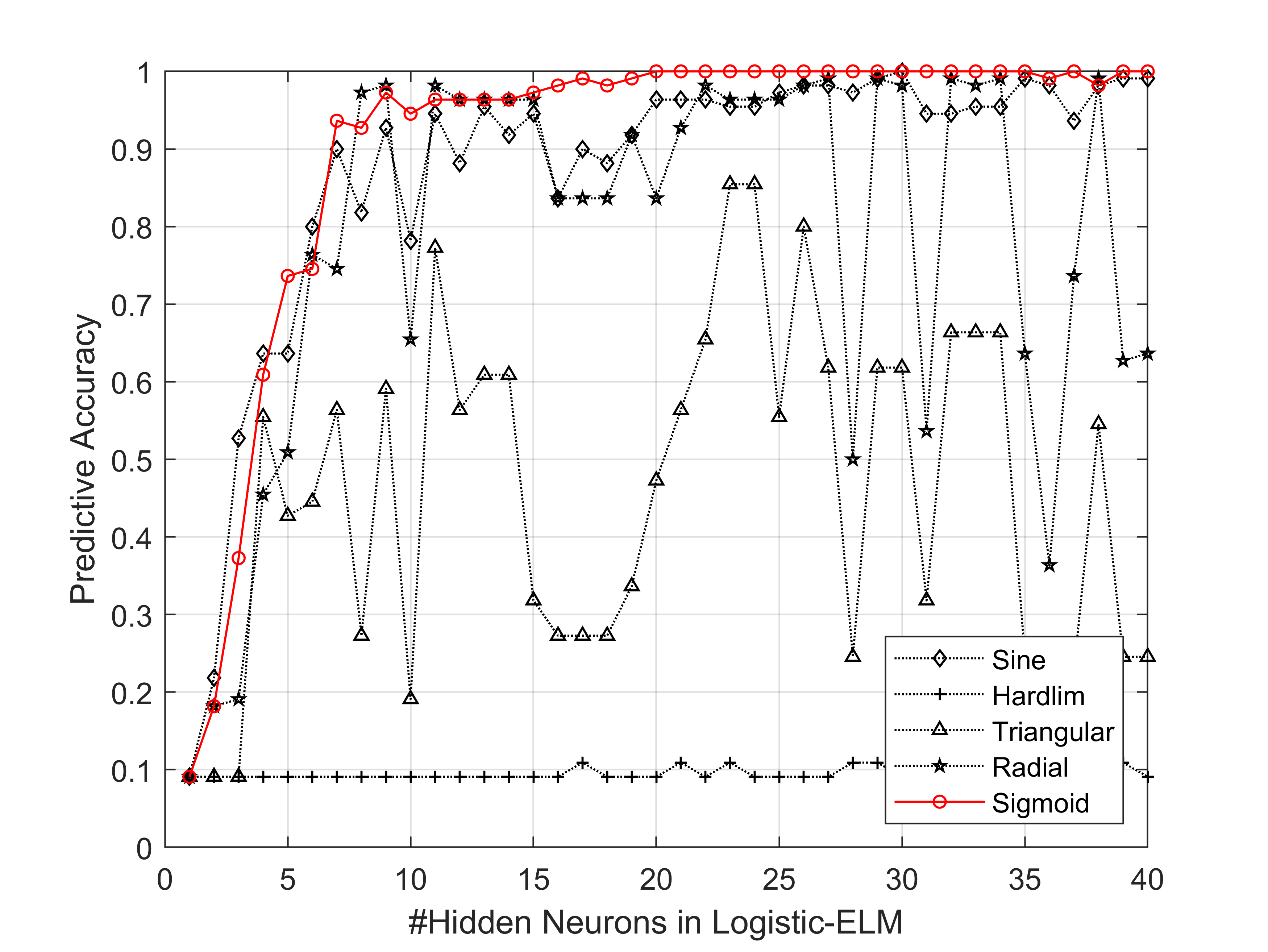}
\caption{The predictive accuracy of logistic-ELM under different activation functions and various number of hidden neurons.}
\label{fig7}
\end{figure}

This experiment is to select the appropriate activation function and the number of hidden neurons for logistic-ELM. We deploy the experiment on CWRU dataset with fault diameter d = 0.007’’\&0.021’’, and Fig.~\ref{fig7} shows the accuracies of logistic-ELM under Sigmoid, Sine, Hardlim, Triangular, and Radial functions, respectively, with various number of hidden neurons.

Overall, the predictive accuracy of Sigmoid is higher than other four activation functions. For Sigmoid, when the number of hidden neurons ranges from 1 to 20, the predictive accuracy increases as neurons number get higher. When the number of hidden neurons ranges from 20 to 30, the predictive accuracy maintains steady, although has little different. When the number of hidden neurons is bigger than 30, the predictive accuracy is 1, but algorithms may over-fit the training set now. 

In order to scientifically describe the above phenomena, Pearson correlation coefficient \cite{39} is used to calculate the linear correlation between prediction accuracy and the number of hidden neurons, as

\begin{equation}
    \rho(accuracy, neuron)=\frac{\sigma_{accuracy,neuron }}{\sigma_{accuracy} \sigma_{neuron}}
\end{equation}

where $\sigma_{accuracy,neuron}$ is the covariance between the predictive accuracy and the number of hidden neurons. $\sigma_{accuracy}$ and $\sigma_{neuron}$ are variances of the predictive accuracy and the number of hidden neurons, respectively. Pearson correlation coefficient returns a value between -1 and 1. A higher absolute value of the correlation coefficient indicates a stronger relationship between variables.

When the number of hidden neurons ranges from 1 to 20, Pearson correlation coefficient between prediction accuracy and the number of hidden neurons is 0.833, which indicates a strong positive correlation. When the number of hidden neurons ranges from 20 to 30, Pearson correlation coefficient is 0.116, which indicates a weak positive correlation. Therefore, we choose Sigmoid as the activation function and 20 as the number of hidden neurons.

\subsection{Experiment 2. Select Initial Chaotic Values for Logistic Mapping. }
\begin{table}[!t]
 \renewcommand{\arraystretch}{1.3}
\caption{Predictive accuracy of logistic-ELM under different initial value of $\mu$ and $z_1$}
\label{tab3}
\centering
\setlength{\tabcolsep}{4.75mm}{
\begin{tabular}{llllll}
\toprule
$z_1$&	\multicolumn{5}{c}{$\mu$}\\
\cline{2-6}
&3.95&	3.96&	3.97&	3.98&	3.99\\
\hline
0.1&	1&	1&	0.98&	0.99&	1\\

0.2&	0.99&	0.99&	0.99&	0.99&	0.99\\

0.3	&1&	1&	0.99&	0.99&	0.99\\

0.4&	1&	0.98&	0.99&	0.98&	0.99\\

0.5&	1&	0.98&	0.98&	0.98&	0.99\\
0.6&	1&	0.99&	0.96&	1&	1\\
0.7&	0.98&	1&	0.99&	1&	1\\
0.8&	0.98&	0.98&	1&	0.96&	1\\
0.9	&0.99&	0.98&	0.99&	0.99&	0.99\\

\hline
\end{tabular}}
\end{table}
This experiment is to select initial chaotic values for logistic-ELM. We deploy the experiment under different initial chaotic value of $z_1$ and $\mu$ generated through logistic mapping. Table~\ref{tab3} shows the predictive accuracies of logistic-ELM on CWRU dataset with fault diameter d = 0.007’’\&0.021’. Logistic mapping is sensitive to initial conditions and the closer $\mu$ is to 4, the more chaotic the system is. Therefore, let the ranges of initial values be $z_1\in [0.1,0.9]$ and $\mu \in[3.95,3.99]$.

As shown in Fig.~\ref{fig4}, when one of $z_1$ and $\mu$ is constant, there is no correlation between the other one and the prediction accuracy. Therefore, without considering the correlation, we choose $z_1$ and $\mu$ with a prediction accuracy of 1 in the table as the initial values of the logical mapping.

\subsection{Experiment 3. Select Appropriate Time-domain Features for Logistic-ELM}
This experiment intends to show the process of feature selection under Sequence Forward Selection (SFS). We execute our scheme with different combination of features by several rounds. 

As shown in Table~\ref{tabfeature}, the initial feature pool is empty in the first round and we run our scheme with each of the 14 features, respectively. Then we record the results and find that Feature 7 makes the highest accuracy 0.79. Therefore, we put Feature 7 in the feature pool. In the second round, we run our scheme with Feature 7 and each of the remaining 13 features. Then we find that Feature 4 makes the highest accuracy 0.92 which is higher than 0.79. So we put Feature 4 in the feature pool. Now the feature pool has two elements. Then we repeat the operations above until the accuracy do not increase anymore. In this way, we obtain the feature pool of \{Feature 7, Feature 4, Feature 9, Feature 6\}.

\begin{table}[!t]
\renewcommand{\arraystretch}{1.3}
\caption{Process of Feature Selection Under SFS.}\label{tabfeature}
\centering

\setlength{\tabcolsep}{2mm}{
\begin{tabular}{cccccc}
\hline
 & Round 1 & Round 2 & Round 3 & Round 4  \\
\hline
Initial Feature Pool & $\emptyset$ & \{F7\} & \{F7,F4\} & \{F7,F4,F9\}  \\
Feature 1 & 0.19 & 0.72 & 0.95 & 0.96 \\
Feature 2 & 0.76 & 0.80 & 0.93 & 0.99  \\
Feature 3 & 0.48 & 0.86 & 0.93 & 0.98 \\
Feature 4 & 0.70 & {\bfseries 0.92} & - & -  \\
Feature 5 & 0.59 & 0.87 & 0.93 & 0.99  \\
Feature 6 & 0.72 & 0.89 & 0.95 & {\bfseries 1.00} \\
Feature 7 & {\bfseries 0.79} & - & - & -  \\
Feature 8 & 0.63 & 0.87 & 0.91 & 0.96  \\
Feature 9 & 0.58 & 0.92 & {\bfseries 0.98} & -  \\
Feature 10 & 0.54 & 0.83 & 0.95 & 0.98  \\
Feature 11 & 0.58 & 0.80 & 0.88 & 0.96  \\
Feature 12 & 0.58 & 0.77 & 0.84 & 0.95  \\
Feature 13 & 0.43 & 0.70 & 0.91 & 0.96  \\
Feature 14 & 0.75 & 0.85 & 0.97 & 0.97  \\
Selected & Feature 7 & Feature 4 & Feature 9 & Feature 6  \\
Feature Pool & \{F7\} & \{F7,F4\} & \{F7,F4,F9\} & {\bfseries \{F7,F4,F9,F6\}} \\
\hline
\end{tabular}}
\end{table}

\subsection{Experiment 4. Predictive Accuracy Comparison with Original ELM}
In this paper, logistic mapping is used to generate the input weights only once, and these weights can be directly used in the subsequent classification. Due to the chaotic nature of logistic mapping, we do not need to set biases while the original ELM randomly generates the input weights and biases to fit the generality. This experiment is to verify the efficiency and stability of logistic-ELM comparing with original ELM.

\begin{figure}[!t]
\centering
\includegraphics[width=3.5in]{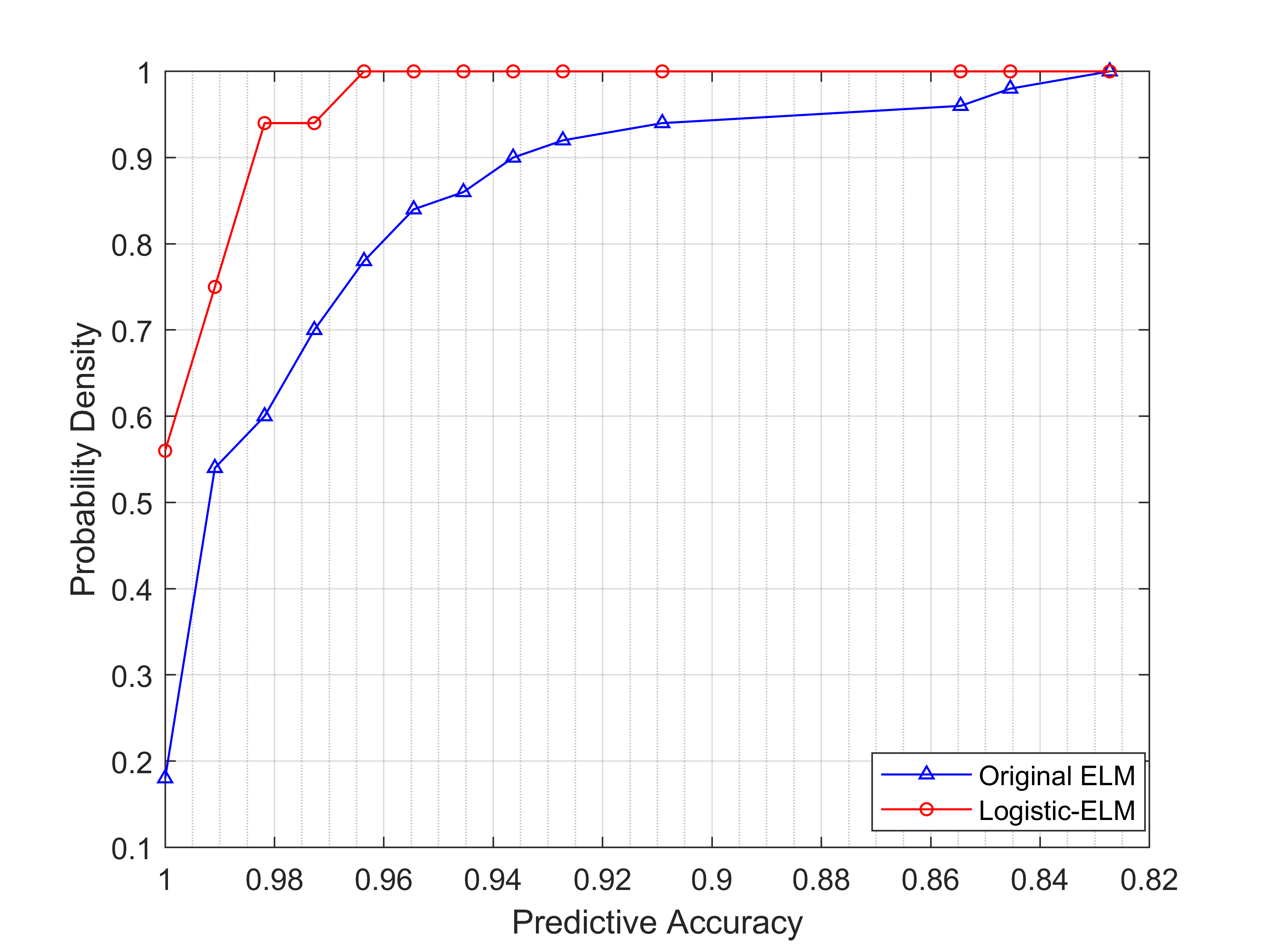}
\caption{Predictive accuracy comparison between original ELM and logistic-ELM.}
\label{fig8}
\end{figure}

 In order to observe the accuracy and stability of logistic-ELM versus the original ELM, we conducted several tests on each of them on the CWRU dataset with fault diameter d = 0.007’’\&0.021’. The input weight matrix $W$ of the logistic-ELM is generated by the logistic mapping using the initial values chosen in Experiment 2 and omitting the bias $b$. In original ELM, $W$ and $b$ are generated randomly. Other parameters are identical for both methods. As shown in table \ref{tab4}, the expected value of the original ELM prediction accuracy is 0.971, which is lower than the expected value of logistic-ELM of 0.993, although the highest prediction accuracy of both methods is 1. Moreover, the accuracy distribution of original ELM is scattered while logistic-ELM is stable. As shown in Fig. \ref{fig8}, the red line indicates logistic-ELM and the blue line indicates the original ELM. The horizontal coordinate of each point in the figure is the prediction accuracy and the vertical coordinate indicates the probability density for each prediction accuracy. Observe that the probability density of logistic-ELM is higher than it of the original ELM at each accuracy. Therefore, logistic-ELM is more stable and has higher prediction accuracy than the original ELM.

\begin{table}[!t]
 \renewcommand{\arraystretch}{1.3}
\caption{Predictive accuracy comparison between original ELM and logistic-ELM}
\label{tab4}
\centering
\setlength{\tabcolsep}{7mm}{
\begin{tabular}{ccc}
\toprule
Prediction &	\multicolumn{2}{c}{Probability Density}\\
\cline{2-3}
Accuracy&Original ELM&	Logistic-ELM\\
\hline
1&	0.18&	0.56\\
$\left[\right.$0.99,1)&	0.36&	0.19\\
$\left[\right.$0.98,0.99)&	0.06&	0.19\\
$\left[\right.$0.97,0.98)&	0.1&	0\\
$\left[\right.$0.96,0.97)&	0.08&	0.06\\
$\left[\right.$0.95,0.96)&	0.06&	0\\
$\left[\right.$0.94,0.95)&0.02&	0\\
$\left[\right.$0.93,0.94)	&0.04&	0\\
$\left[\right.$0.92,0.93)&	0&	0\\
$\left[\right.$0.91,0.92)&0.02&	0\\
$\left[\right.$0.90,0.91)&	0&	0\\
$\left[\right.$0,0.90)&	0.06&	0\\
Highest	&1&	1\\
Expected&	0.971	&{\bfseries 0.993}\\
\hline
\end{tabular}}
\end{table}

\subsection{Experiment 5. Predictive Accuracy under Different Operating Conditions}
The above-mentioned experiments demonstrate that the logistic-ELM is a fast diagnosis method with high predictive accuracy, but are all deployed under 0 hp motor load. In order to verify the logistic-ELM’s adaptability to different operating conditions, we perform experiment to observe the predictive accuracy of fault diagnosis of the rolling bearings under four different loads (0 hp, 1hp, 2 hp, and 3 hp) selected from CWRU dataset. Table \ref{tab5} shows the results under different operating conditions. We find that the highest predictive accuracy of logistic-ELM is 100\% in these conditions. It indicates that the logistic-ELM can adapt to variable operating conditions.

\begin{table}[!t]
 \renewcommand{\arraystretch}{1.3}
\caption{Predictive accuracy comparison between original ELM and logistic-ELM}
\label{tab5}
\centering
\setlength{\tabcolsep}{5mm}{
\begin{tabular}{llll}
\toprule
Condition&	Speed(r/min)&	Load(hp)&	Accuracy\\
\hline
Condition 1	&1797&	0&	100\% \\
Condition 2	&1772&	1&	100\% \\
Condition 3	&1750&	2&	100\% \\
Condition 4&	1730&	3&	100\% \\
Average		&&&	100\% \\

\hline
\end{tabular}}
\end{table}

\subsubsection{Experiment 6. Predictive Accuracy Comparisons with Existing Fault Diagnosis Methods}
In this experiment, we select four existing fault diagnosis methods of rolling bearings for comparisons with proposed logistic-ELM, including DSLS-SVM \cite{33}, PMSEn \cite{12}, FuzzyMEn \cite{13}, VAEGAN-DRA \cite{34}, RCFOA-ELM\cite{37} and DE-ELM \cite{38}, which were proposed from 2019 to 2021. Among these methods, DSLS-SVM is an extended algorithm of the SVM. PMSEn and FuzzyMEn are extended algorithms of the multi-scale entropy, while VAEGAN-DRA utilizes the deep learning method GAN. RCFOA-ELM and DE-ELM are based on ELM. Datasets in these methods are subsets of CWRU but different, so we perform each comparison experiment on the same subset as described in the comparison object, with 5-fold cross-validation. Table \ref{tab6} shows the details of related sub datasets. 

\begin{figure}[!t]
\centering
\includegraphics[width=4in]{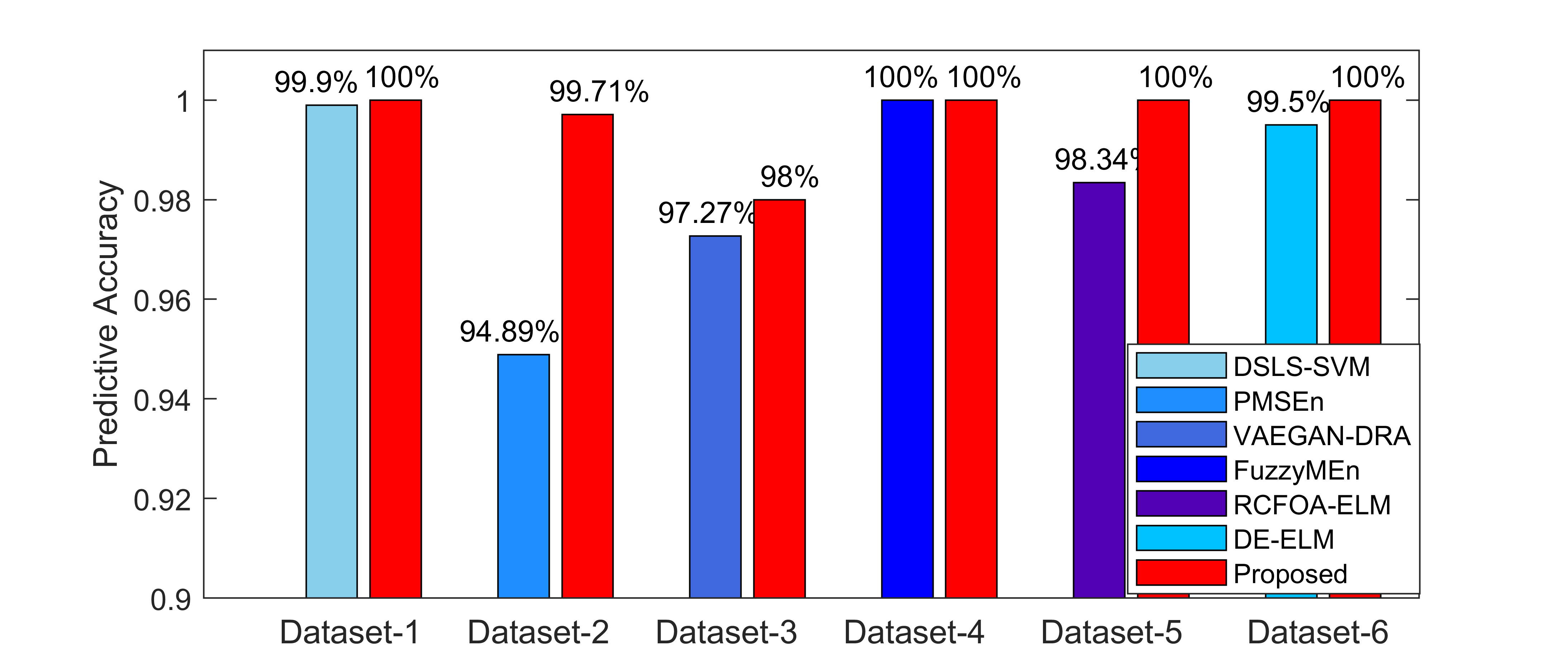}
\caption{Predictive accuracy comparisons with existing fault diagnosis methods.}
\label{fig9}
\end{figure}

Due to the lack of understanding of the implementation details of these methods, the accuracies of related comparison objects are from their public reports. Fig. \ref{fig9} shows the comparisons with predictive accuracy of logistic-ELM. On the four sub datasets, our proposed logistic-ELM obtains accuracies with 100\%, 99.71\%, 98\%, 100\%, 100\%, 100\%, respectively, while those of the four comparison objects are 99.9\%, 94.89\%, 97.27\%, 100\%, 98.34\% and 99.5\%. Results indicate the proposed logistic-ELM is accurate in fault diagnosis and has superiority to existing methods.

\begin{table}[!t]
\caption{Predictive accuracy comparison between original ELM and logistic-ELM}
\label{tab6}
\centering

\begin{tabular}{ll}
\toprule
Sub Dataset&	Description of Fault Information\\
\hline
Dataset-1 in DSLS-SVM \cite{33}&	N(0 hp), IF-BF-OF(0 hp, 0.007’’) \\&from CWRU dataset\\
Dataset-2 in  PMSEn \cite{12}&	N(0 hp), N(1 hp), N(2 hp), N(3 hp), \\&IF-BF-OF(0 hp, 0.021’’) from CWRU dataset\\
Dataset-3 in VAEGAN-DRA \cite{34}&	N (0 hp), IF-BF-OF(0 hp, 0.007’’), \\& IF-BF-OF(0 hp, 0.014’’), IF-BF-OF(0 hp, 0.021’’)\\& from CWRU dataset\\
Dataset-4 in FuzzyMEn \cite{13}&	N(0 hp) , IF-BF-OF(0 hp, 0.007’’), \\&IF-BF(0 hp, 0.021’’), OF(0 hp, 0.028’’)\\& from CWRU dataset\\
Dataset-5 in RCFOA-ELM \cite{37}&	N(2 hp), IF-BF-OF(0 hp, 0.014’’) \\&from CWRU dataset\\
Dataset-6 in DE-ELM \cite{38}&	N(2 hp), IF-BF-OF(2 hp, 0.007’’)\\& from CWRU dataset\\
\hline
\end{tabular}

\end{table}

\subsection{Experiment 7. Runtime Cost Comparisons with Existing Fault Diagnosis Methods}
This experiment is to test the runtime cost of logistic-ELM by comparing with existing fault diagnosis methods. We select DSLS-LSSVM \cite{33}, PMSEn \cite{12} and FS-CNN \cite{35} for comparisons, which are based on SVM, multi-scale entropy, and CNN, on CWRU dataset with fault diameter d = 0.007’’. All the codes are implemented in MATLAB R2018b, in operating system Win10 (64-bit), with a 16G Memory and Intel (R) Core (TM) i7-10875H CPU. To obtain the calculation time in a reasonable manner, the experiments are performed 5 times, and the average time is determined. The dataset for each experiment includes 300 samples, and the length of each sample is 2048. Table \ref{tab7} shows the results, where runtime is just testing time, not training time. As we can see, the average predicting runtime cost is 0.0393 s, while the other three methods need 0.8311 s, 73.04 s, 3.586 s, respectively. Compared with the existing method, the operating efficiency of logistic-ELM is increased by 21.15 times, 1858.52 times and 9.12 times respectively. Results indicate the proposed logistic-ELM is a very fast fault diagnosis method for rolling bearings.

\begin{table}[!t]
 \renewcommand{\arraystretch}{1.3}
\caption{Runtime Cost Comparisons (s)}
\label{tab7}
\centering
\setlength{\tabcolsep}{7mm}{
\begin{tabular}{cc}
\toprule
Method&	Average Runtime Cost (s)\\
\hline
DSLS-LSSVM&	0.8311\\
PMSEn&	73.04\\
FS-CNN&	3.586\\
Proposed&	0.0393\\

\hline
\end{tabular}}
\end{table}

\section{Conclusion}
\label{section5}
This paper proposes a fast and efficient fault diagnosis method for rolling bearings, named logistic-ELM. We first extract a group of optimal features to build the feature extractor by SFS from 14 time-domain feature models based on original vibration signals. Then we utilize ELM as the baseline classifier for fault diagnosis but replace the dynamic random input weights with pseudo-random sequence generated by chaotic logistic mapping. We perform a series of experiments based on CWRU dataset, results indicate that the proposed logistic-ELM has great high predictive accuracy, very short runtime, and can adapt to different operating conditions. During all comparisons, the logistic-ELM presents high superiority in fault diagnosis of rolling bearings. In future work, we will try the semi-supervised or unsupervised learning models to reduce the dependence on datasets with annotation errors.

\section*{Contribution Authors}
Zhenhua Tan proposes the original opinion, methodology, design, original architecture and writing. Jingyu Ning is responsible for the design, coding, data analysis, original writing and revision. Kai peng participates the coding, data analysis and revision, while Zhenche Xia and Danke Wu participates the design and revision.

\section*{Acknowledgment}

This work was funded by the National Key Research and Development Program of China under Grant No. 2019YFB1405803.


\bibliography{sn-bibliography}


\begin{thebibliography}{41}
\ifx \bisbn   \undefined \def \bisbn  #1{ISBN #1}\fi
\ifx \binits  \undefined \def \binits#1{#1}\fi
\ifx \bauthor  \undefined \def \bauthor#1{#1}\fi
\ifx \batitle  \undefined \def \batitle#1{#1}\fi
\ifx \bjtitle  \undefined \def \bjtitle#1{#1}\fi
\ifx \bvolume  \undefined \def \bvolume#1{\textbf{#1}}\fi
\ifx \byear  \undefined \def \byear#1{#1}\fi
\ifx \bissue  \undefined \def \bissue#1{#1}\fi
\ifx \bfpage  \undefined \def \bfpage#1{#1}\fi
\ifx \blpage  \undefined \def \blpage #1{#1}\fi
\ifx \burl  \undefined \def \burl#1{\textsf{#1}}\fi
\ifx \doiurl  \undefined \def \doiurl#1{\url{https://doi.org/#1}}\fi
\ifx \betal  \undefined \def \betal{\textit{et al.}}\fi
\ifx \binstitute  \undefined \def \binstitute#1{#1}\fi
\ifx \binstitutionaled  \undefined \def \binstitutionaled#1{#1}\fi
\ifx \bctitle  \undefined \def \bctitle#1{#1}\fi
\ifx \beditor  \undefined \def \beditor#1{#1}\fi
\ifx \bpublisher  \undefined \def \bpublisher#1{#1}\fi
\ifx \bbtitle  \undefined \def \bbtitle#1{#1}\fi
\ifx \bedition  \undefined \def \bedition#1{#1}\fi
\ifx \bseriesno  \undefined \def \bseriesno#1{#1}\fi
\ifx \blocation  \undefined \def \blocation#1{#1}\fi
\ifx \bsertitle  \undefined \def \bsertitle#1{#1}\fi
\ifx \bsnm \undefined \def \bsnm#1{#1}\fi
\ifx \bsuffix \undefined \def \bsuffix#1{#1}\fi
\ifx \bparticle \undefined \def \bparticle#1{#1}\fi
\ifx \barticle \undefined \def \barticle#1{#1}\fi
\bibcommenthead
\ifx \bconfdate \undefined \def \bconfdate #1{#1}\fi
\ifx \botherref \undefined \def \botherref #1{#1}\fi
\ifx \url \undefined \def \url#1{\textsf{#1}}\fi
\ifx \bchapter \undefined \def \bchapter#1{#1}\fi
\ifx \bbook \undefined \def \bbook#1{#1}\fi
\ifx \bcomment \undefined \def \bcomment#1{#1}\fi
\ifx \oauthor \undefined \def \oauthor#1{#1}\fi
\ifx \citeauthoryear \undefined \def \citeauthoryear#1{#1}\fi
\ifx \endbibitem  \undefined \def \endbibitem {}\fi
\ifx \bconflocation  \undefined \def \bconflocation#1{#1}\fi
\ifx \arxivurl  \undefined \def \arxivurl#1{\textsf{#1}}\fi
\csname PreBibitemsHook\endcsname

\bibitem{1}
\begin{bchapter}
\bauthor{\bsnm{Sheng}, \binits{S.}}:
\bctitle{Wind turbine gearbox reliability database, condition monitoring}.
(\byear{2016}).
\burl{https://www.nrel.gov/docs/fy16osti/66028.pdf}
\end{bchapter}
\endbibitem

\bibitem{2}
\begin{botherref}
\oauthor{\bsnm{Wheeler}, \binits{P.G.}}:
Bearing analysis keeps downtime down.
Plant Engineering
(25),
87--89
(1968)
\end{botherref}
\endbibitem

\bibitem{3}
\begin{botherref}
\oauthor{\bsnm{Martin}, \binits{R.L.}}:
Detection of ball bearing malfunctions
(1970)
\end{botherref}
\endbibitem

\bibitem{4}
\begin{botherref}
\oauthor{\bsnm{Boto}, \binits{P.A.}}:
Detection of bearing damage by shock pulse measurement.
Ball Bearing Journal
(1971)
\end{botherref}
\endbibitem

\bibitem{5}
\begin{botherref}
\oauthor{\bsnm{Harting}, \binits{D.R.}},
\oauthor{\bsnm{Taylor}, \binits{J.W.}}:
Demodulated resonance analysis system.
US
(1974)
\end{botherref}
\endbibitem

\bibitem{6}
\begin{barticle}
\bauthor{\bsnm{Cooley}, \binits{J.W.}},
\bauthor{\bsnm{Tukey}, \binits{J.W.}}:
\batitle{{An Algorithm for the Machine Calculation of Complex Fourier Series}}.
\bjtitle{Math. Comput.}
\bvolume{19},
\bfpage{297}--\blpage{301}
(\byear{1965}).
\doiurl{10.1090/S0025-5718-1965-0178586-1}
\end{barticle}
\endbibitem

\bibitem{7}
\begin{barticle}
\bauthor{\bsnm{Nikula}, \binits{R.-P.}},
\bauthor{\bsnm{Karioja}, \binits{K.}},
\bauthor{\bsnm{Pylvänäinen}, \binits{M.}},
\bauthor{\bsnm{Leiviskä}, \binits{K.}}:
\batitle{Automation of low-speed bearing fault diagnosis based on
  autocorrelation of time domain features}.
\bjtitle{Mechanical Systems and Signal Processing}
\bvolume{138},
\bfpage{106572}
(\byear{2020}).
\doiurl{10.1016/j.ymssp.2019.106572}
\end{barticle}
\endbibitem

\bibitem{8}
\begin{barticle}
\bauthor{\bsnm{Richman}, \binits{J.S.}},
\bauthor{\bsnm{Moorman}, \binits{J.R.}}:
\batitle{Physiological time-series analysis using approximate entropy and
  sample entropy}.
\bjtitle{American Journal of Physiology-Heart and Circulatory Physiology}
\bvolume{278}(\bissue{6}),
\bfpage{2039}--\blpage{2049}
(\byear{2000}).
\doiurl{10.1152/ajpheart.2000.278.6.H2039}.
\bcomment{PMID: 10843903}
\end{barticle}
\endbibitem

\bibitem{9}
\begin{barticle}
\bauthor{\bsnm{Pincus}, \binits{S.M.}}:
\batitle{Approximate entropy as a measure of system complexity.}
\bjtitle{Proceedings of the National Academy of Sciences}
\bvolume{88}(\bissue{6}),
\bfpage{2297}--\blpage{2301}
(\byear{1991}).
\doiurl{10.1073/pnas.88.6.2297}
\end{barticle}
\endbibitem

\bibitem{10}
\begin{barticle}
\bauthor{\bsnm{Li}, \binits{X.}},
\bauthor{\bsnm{Cui}, \binits{S.}},
\bauthor{\bsnm{Voss}, \binits{L.}}:
\batitle{{Using Permutation Entropy to Measure the Electroencephalographic
  Effects of Sevoflurane}}.
\bjtitle{Anesthesiology}
\bvolume{109}(\bissue{3}),
\bfpage{448}--\blpage{456}
(\byear{2008}).
\doiurl{10.1097/ALN.0b013e318182a91b}
\end{barticle}
\endbibitem

\bibitem{11}
\begin{barticle}
\bauthor{\bsnm{Nicolaou}, \binits{N.}},
\bauthor{\bsnm{Georgiou}, \binits{J.}}:
\batitle{Detection of epileptic electroencephalogram based on permutation
  entropy and support vector machines}.
\bjtitle{Expert Systems with Applications}
\bvolume{39}(\bissue{1}),
\bfpage{202}--\blpage{209}
(\byear{2012}).
\doiurl{10.1016/j.eswa.2011.07.008}
\end{barticle}
\endbibitem

\bibitem{12}
\begin{barticle}
\bauthor{\bsnm{{Zhao}}, \binits{D.}},
\bauthor{\bsnm{{Liu}}, \binits{S.}},
\bauthor{\bsnm{{Cheng}}, \binits{S.}},
\bauthor{\bsnm{{Sun}}, \binits{X.}},
\bauthor{\bsnm{{Wang}}, \binits{L.}},
\bauthor{\bsnm{{Wei}}, \binits{Y.}},
\bauthor{\bsnm{{Zhang}}, \binits{H.}}:
\batitle{{Parallel multi-scale entropy and it's application in rolling bearing
  fault diagnosis}}.
\bjtitle{Measurements}
\bvolume{168},
\bfpage{108333}
(\byear{2021}).
\doiurl{10.1016/j.measurement.2020.108333}
\end{barticle}
\endbibitem

\bibitem{13}
\begin{bchapter}
\bauthor{\bsnm{Howedi}, \binits{A.}},
\bauthor{\bsnm{Lotfi}, \binits{A.}},
\bauthor{\bsnm{Pourabdollah}, \binits{A.}}:
\bctitle{A multi-scale fuzzy entropy measure for anomaly detection in
  activities of daily living}.
In: \bbtitle{Proceedings of the 13th ACM International Conference on PErvasive
  Technologies Related to Assistive Environments}.
\bsertitle{PETRA '20}.
\bpublisher{Association for Computing Machinery},
\blocation{New York, NY, USA}
(\byear{2020}).
\doiurl{10.1145/3389189.3397987}
\end{bchapter}
\endbibitem

\bibitem{14}
\begin{barticle}
\bauthor{\bsnm{Stack}, \binits{J.R.}},
\bauthor{\bsnm{Harley}, \binits{R.G.}},
\bauthor{\bsnm{Habetler}, \binits{T.G.}}:
\batitle{An amplitude modulation detector for fault diagnosis in rolling
  element bearings}.
\bjtitle{IEEE Transactions on Industrial Electronics}
\bvolume{51},
\bfpage{1097}--\blpage{1102}
(\byear{2004})
\end{barticle}
\endbibitem

\bibitem{15}
\begin{barticle}
\bauthor{\bsnm{Blankenship}, \binits{G.W.}},
\bauthor{\bsnm{Singh}, \binits{R.}}:
\batitle{Analytical solution for modulation sidebands associated with a class
  of mechanical oscillators}.
\bjtitle{Journal of Sound and Vibration}
\bvolume{179}(\bissue{1}),
\bfpage{13}--\blpage{36}
(\byear{1995}).
\doiurl{10.1006/jsvi.1995.0002}
\end{barticle}
\endbibitem

\bibitem{16}
\begin{barticle}
\bauthor{\bsnm{Liang}, \binits{B.}},
\bauthor{\bsnm{Iwnicki}, \binits{S.D.}},
\bauthor{\bsnm{Zhao}, \binits{Y.}}:
\batitle{Application of power spectrum, cepstrum, higher order spectrum and
  neural network analyses for induction motor fault diagnosis}.
\bjtitle{Mechanical Systems and Signal Processing}
\bvolume{39}(\bissue{1}),
\bfpage{342}--\blpage{360}
(\byear{2013}).
\doiurl{10.1016/j.ymssp.2013.02.016}
\end{barticle}
\endbibitem

\bibitem{17}
\begin{barticle}
\bauthor{\bsnm{Feng}, \binits{Z.}},
\bauthor{\bsnm{Liang}, \binits{M.}},
\bauthor{\bsnm{Chu}, \binits{F.}}:
\batitle{Recent advances in time–frequency analysis methods for machinery
  fault diagnosis: A review with application examples}.
\bjtitle{Mechanical Systems and Signal Processing}
\bvolume{38}(\bissue{1}),
\bfpage{165}--\blpage{205}
(\byear{2013}).
\doiurl{10.1016/j.ymssp.2013.01.017}.
\bcomment{Condition monitoring of machines in non-stationary operations.}
\end{barticle}
\endbibitem

\bibitem{18}
\begin{barticle}
\bauthor{\bsnm{Jr}, \binits{B.}},
\bauthor{\bsnm{Ware}, \binits{H.A.}},
\bauthor{\bsnm{Wipf}, \binits{D.P.}},
\bauthor{\bsnm{Tompkins}, \binits{W.R.}},
\bauthor{\bsnm{Clark}, \binits{B.R.}},
\bauthor{\bsnm{Larson}, \binits{E.C.}},
\bauthor{\bsnm{Poor}, \binits{H.V.}}:
\batitle{Fault diagnostics using statistical change detection in the bispectral
  domain}.
\bjtitle{Mechanical Systems and Signal Processing}
\bvolume{14}(\bissue{4}),
\bfpage{561}--\blpage{570}
(\byear{2000}).
\doiurl{10.1006/mssp.2000.1299}
\end{barticle}
\endbibitem

\bibitem{19}
\begin{barticle}
\bauthor{\bsnm{Tian}, \binits{J.}},
\bauthor{\bsnm{Morillo}, \binits{C.}},
\bauthor{\bsnm{Azarian}, \binits{M.}},
\bauthor{\bsnm{Pecht}, \binits{M.}}:
\batitle{Motor bearing fault detection using spectral kurtosis based feature
  extraction and k-nearest neighbor distance analysis}.
\bjtitle{IEEE Transactions on Industrial Electronics}
\bvolume{63},
\bfpage{1}--\blpage{1}
(\byear{2015}).
\doiurl{10.1109/TIE.2015.2509913}
\end{barticle}
\endbibitem

\bibitem{20}
\begin{barticle}
\bauthor{\bsnm{Antoni}, \binits{J.}},
\bauthor{\bsnm{Randall}, \binits{R.B.}}:
\batitle{The spectral kurtosis: application to the vibratory surveillance and
  diagnostics of rotating machines}.
\bjtitle{Mechanical Systems and Signal Processing}
\bvolume{20}(\bissue{2}),
\bfpage{308}--\blpage{331}
(\byear{2006}).
\doiurl{10.1016/j.ymssp.2004.09.002}
\end{barticle}
\endbibitem

\bibitem{21}
\begin{barticle}
\bauthor{\bsnm{{Huang}}, \binits{N.E.}},
\bauthor{\bsnm{{Shen}}, \binits{Z.}},
\bauthor{\bsnm{{Long}}, \binits{S.R.}},
\bauthor{\bsnm{{Wu}}, \binits{M.C.}},
\bauthor{\bsnm{{Shih}}, \binits{H.H.}},
\bauthor{\bsnm{{Zheng}}, \binits{Q.}},
\bauthor{\bsnm{{Yen}}, \binits{N.-C.}},
\bauthor{\bsnm{{Tung}}, \binits{C.C.}},
\bauthor{\bsnm{{Liu}}, \binits{H.H.}}:
\batitle{{The empirical mode decomposition and the Hilbert spectrum for
  nonlinear and non-stationary time series analysis}}.
\bjtitle{Proceedings of the Royal Society of London Series A}
\bvolume{454}(\bissue{1971}),
\bfpage{903}--\blpage{998}
(\byear{1998}).
\doiurl{10.1098/rspa.1998.0193}
\end{barticle}
\endbibitem

\bibitem{22}
\begin{barticle}
\bauthor{\bsnm{Smith}, \binits{J.R.}}:
\batitle{The local mean decomposition and its application to eeg perception
  data}.
\bjtitle{Journal of The Royal Society Interface}
\bvolume{2},
\bfpage{443}--\blpage{454}
(\byear{2005})
\end{barticle}
\endbibitem

\bibitem{23}
\begin{bchapter}
\bauthor{\bsnm{Vapnik}, \binits{V.N.}}:
\bctitle{The nature of statistical learning theory}.
In: \bbtitle{Statistics for Engineering and Information Science}
(\byear{2000})
\end{bchapter}
\endbibitem

\bibitem{24}
\begin{barticle}
\bauthor{\bsnm{Konar}, \binits{P.}},
\bauthor{\bsnm{Chattopadhyay}, \binits{P.}}:
\batitle{Bearing fault detection of induction motor using wavelet and support
  vector machines (svms)}.
\bjtitle{Applied Soft Computing}
\bvolume{11}(\bissue{6}),
\bfpage{4203}--\blpage{4211}
(\byear{2011}).
\doiurl{10.1016/j.asoc.2011.03.014}
\end{barticle}
\endbibitem

\bibitem{25}
\begin{barticle}
\bauthor{\bsnm{Muralidharan}, \binits{V.}},
\bauthor{\bsnm{Sugumaran}, \binits{V.}}:
\batitle{A comparative study of naïve bayes classifier and bayes net
  classifier for fault diagnosis of monoblock centrifugal pump using wavelet
  analysis}.
\bjtitle{Applied Soft Computing}
\bvolume{12}(\bissue{8}),
\bfpage{2023}--\blpage{2029}
(\byear{2012}).
\doiurl{10.1016/j.asoc.2012.03.021}
\end{barticle}
\endbibitem

\bibitem{27}
\begin{botherref}
\oauthor{\bsnm{Lecun}, \binits{Y.}},
\oauthor{\bsnm{Bengio}, \binits{Y.}}:
Convolutional networks for images, speech, and time-series
(1995)
\end{botherref}
\endbibitem

\bibitem{28}
\begin{bchapter}
\bauthor{\bsnm{Huang}, \binits{G.-B.}},
\bauthor{\bsnm{Zhu}, \binits{Q.-Y.}},
\bauthor{\bsnm{Siew}, \binits{C.-K.}}:
\bctitle{Extreme learning machine: a new learning scheme of feedforward neural
  networks}.
In: \bbtitle{2004 IEEE International Joint Conference on Neural Networks (IEEE
  Cat. No.04CH37541)},
vol. \bseriesno{2},
pp. \bfpage{985}--\blpage{9902}
(\byear{2004}).
\doiurl{10.1109/IJCNN.2004.1380068}
\end{bchapter}
\endbibitem

\bibitem{29}
\begin{bchapter}
\bauthor{\bsnm{Kuspijani}, \binits{K.}},
\bauthor{\bsnm{Watiasih}, \binits{R.}},
\bauthor{\bsnm{Prihastono}, \binits{P.}}:
\bctitle{Faults identification of induction motor based on vibration using
  backpropagation neural network}.
In: \bbtitle{2020 International Conference on Smart Technology and Applications
  (ICoSTA)},
pp. \bfpage{1}--\blpage{5}
(\byear{2020}).
\doiurl{10.1109/ICoSTA48221.2020.1570615779}
\end{bchapter}
\endbibitem

\bibitem{30}
\begin{bchapter}
\bauthor{\bsnm{Gao}, \binits{D.}},
\bauthor{\bsnm{Zhu}, \binits{Y.}},
\bauthor{\bsnm{Wang}, \binits{X.}},
\bauthor{\bsnm{Yan}, \binits{K.}},
\bauthor{\bsnm{Hong}, \binits{J.}}:
\bctitle{A fault diagnosis method of rolling bearing based on complex morlet
  cwt and cnn}.
In: \bbtitle{2018 Prognostics and System Health Management Conference
  (PHM-Chongqing)},
pp. \bfpage{1101}--\blpage{1105}
(\byear{2018}).
\doiurl{10.1109/PHM-Chongqing.2018.00194}
\end{bchapter}
\endbibitem

\bibitem{37}
\begin{barticle}
\bauthor{\bsnm{He}, \binits{C.}},
\bauthor{\bsnm{Wu}, \binits{T.}},
\bauthor{\bsnm{Gu}, \binits{R.}},
\bauthor{\bsnm{Jin}, \binits{Z.}},
\bauthor{\bsnm{Ma}, \binits{R.}},
\bauthor{\bsnm{Qu}, \binits{H.}}:
\batitle{Rolling bearing fault diagnosis based on composite multiscale
  permutation entropy and reverse cognitive fruit fly optimization algorithm
  – extreme learning machine}.
\bjtitle{Measurement}
\bvolume{173},
\bfpage{108636}
(\byear{2021}).
\doiurl{10.1016/j.measurement.2020.108636}
\end{barticle}
\endbibitem

\bibitem{38}
\begin{barticle}
\bauthor{\bsnm{Hu}, \binits{Y.}},
\bauthor{\bsnm{Dong}, \binits{M.}},
\bauthor{\bsnm{Wang}, \binits{G.}},
\bauthor{\bsnm{Fan}, \binits{Z.}},
\bauthor{\bsnm{Zhang}, \binits{S.}}:
\batitle{New method of bearing fault diagnosis based on mmemd and de\_elm}.
\bjtitle{The Journal of Engineering}
\bvolume{2019}(\bissue{23}),
\bfpage{9152}--\blpage{9156}
(\byear{2019}).
\doiurl{10.1049/joe.2018.9206}
\end{barticle}
\endbibitem

\bibitem{36}
\begin{barticle}
\bauthor{\bsnm{May}, \binits{R.}}:
\batitle{Simple mathematical models with very complicated dynamics}.
\bjtitle{Nature}
\bvolume{26},
\bfpage{457}
(\byear{1976}).
\doiurl{10.1038/261459a0}
\end{barticle}
\endbibitem

\bibitem{41}
\begin{botherref}
\oauthor{\bsnm{Li}, \binits{Y.}},
\oauthor{\bsnm{Gao}, \binits{Q.}},
\oauthor{\bsnm{Miao}, \binits{B.}},
\oauthor{\bsnm{Zhang}, \binits{W.}},
\oauthor{\bsnm{Liu}, \binits{J.}},
\oauthor{\bsnm{Zhu}, \binits{Y.}}:
Application of the refined multiscale permutation entropy method to fault
  detection of rolling bearing.
Journal of the Brazilian Society of Mechanical Sciences and Engineering
\textbf{43}
(2021).
\doiurl{10.1007/s40430-021-02986-7}
\end{botherref}
\endbibitem

\bibitem{26}
\begin{barticle}
\bauthor{\bsnm{{Rumelhart}}, \binits{D.E.}},
\bauthor{\bsnm{{Hinton}}, \binits{G.E.}},
\bauthor{\bsnm{{Williams}}, \binits{R.J.}}:
\batitle{{Learning representations by back-propagating errors}}.
\bjtitle{Nature}
\bvolume{323}(\bissue{6088}),
\bfpage{533}--\blpage{536}
(\byear{1986}).
\doiurl{10.1038/323533a0}
\end{barticle}
\endbibitem

\bibitem{40}
\begin{botherref}
\oauthor{\bsnm{Han}, \binits{T.}},
\oauthor{\bsnm{Tian}, \binits{Z.}},
\oauthor{\bsnm{Yin}, \binits{Z.}},
\oauthor{\bsnm{Tan}, \binits{A.}}:
Bearing fault identification based on convolutional neural network by different
  input modes.
Journal of the Brazilian Society of Mechanical Sciences and Engineering
\textbf{42}
(2020).
\doiurl{10.1007/s40430-020-02561-6}
\end{botherref}
\endbibitem

\bibitem{31}
\begin{barticle}
\bauthor{\bsnm{Luo}, \binits{M.}},
\bauthor{\bsnm{Li}, \binits{C.}},
\bauthor{\bsnm{Zhang}, \binits{X.}},
\bauthor{\bsnm{Li}, \binits{R.}},
\bauthor{\bsnm{An}, \binits{X.}}:
\batitle{Compound feature selection and parameter optimization of elm for fault
  diagnosis of rolling element bearings}.
\bjtitle{ISA Transactions}
\bvolume{65},
\bfpage{556}--\blpage{566}
(\byear{2016}).
\doiurl{10.1016/j.isatra.2016.08.022}
\end{barticle}
\endbibitem

\bibitem{32}
\begin{bchapter}
\bauthor{\bsnm{KA}, \binits{L.}}:
\bctitle{Case western reserve university bearing data center}.
(\byear{2012}).
\burl{https://engineering.case.edu/bearingdatacenter/download-data-file}
\end{bchapter}
\endbibitem

\bibitem{39}
\begin{botherref}
\oauthor{\bsnm{Profillidis}, \binits{V.A.}},
\oauthor{\bsnm{Botzoris}, \binits{G.N.}}:
Statistical methods for transport demand modeling.
Modeling of Transport Demand,
163--224
(2019).
\doiurl{10.1016/B978-0-12-811513-8.00005-4}
\end{botherref}
\endbibitem

\bibitem{33}
\begin{barticle}
\bauthor{\bsnm{Li}, \binits{X.}},
\bauthor{\bsnm{Yang}, \binits{Y.}},
\bauthor{\bsnm{Pan}, \binits{H.}},
\bauthor{\bsnm{Cheng}, \binits{J.}},
\bauthor{\bsnm{Cheng}, \binits{J.}}:
\batitle{A novel deep stacking least squares support vector machine for rolling
  bearing fault diagnosis}.
\bjtitle{Computers in Industry}
\bvolume{110},
\bfpage{36}--\blpage{47}
(\byear{2019}).
\doiurl{10.1016/j.compind.2019.05.005}
\end{barticle}
\endbibitem

\bibitem{34}
\begin{barticle}
\bauthor{\bsnm{Liu}, \binits{S.}},
\bauthor{\bsnm{Jiang}, \binits{H.}},
\bauthor{\bsnm{Wu}, \binits{Z.}},
\bauthor{\bsnm{Li}, \binits{X.}}:
\batitle{Rolling bearing fault diagnosis using variational autoencoding
  generative adversarial networks with deep regret analysis}.
\bjtitle{Measurement}
\bvolume{168},
\bfpage{108371}
(\byear{2021}).
\doiurl{10.1016/j.measurement.2020.108371}
\end{barticle}
\endbibitem

\bibitem{35}
\begin{bchapter}
\bauthor{\bsnm{Liang}, \binits{P.}},
\bauthor{\bsnm{Deng}, \binits{C.}},
\bauthor{\bsnm{Wu}, \binits{J.}},
\bauthor{\bsnm{Yang}, \binits{Z.}},
\bauthor{\bsnm{Zhu}, \binits{J.}}:
\bctitle{Intelligent fault diagnosis of rolling element bearing based on
  convolutional neural network and frequency spectrograms}.
In: \bbtitle{2019 IEEE International Conference on Prognostics and Health
  Management (ICPHM)},
pp. \bfpage{1}--\blpage{5}
(\byear{2019}).
\doiurl{10.1109/ICPHM.2019.8819444}
\end{bchapter}
\endbibitem

\end{thebibliography}


\end{document}